\documentclass{article}

\usepackage{microtype}
\usepackage{graphicx}
\usepackage{subcaption}
\usepackage{booktabs} 

\usepackage[dvipsnames]{xcolor}
\usepackage{hyperref}

\usepackage{xspace}

\usepackage[accepted]{icml2024}

\usepackage{amsmath}
\usepackage{amssymb}
\usepackage{mathtools}
\usepackage{amsthm}

\usepackage[capitalize,noabbrev]{cleveref}
\usepackage{titlesec}
\titlespacing\section{0pt}{4pt}{0pt}
\titlespacing\subsection{0pt}{2pt}{0pt}
\titlespacing\paragraph{0pt}{0pt}{4pt}

\theoremstyle{plain}

\theoremstyle{definition}

\theoremstyle{remark}

\newcommand{\algname}{\textsc{DAO}\xspace}

\usepackage[textsize=tiny,textwidth=25mm]{todonotes}

\newif\iffinal

\iffinal
    \newcommand{\YC}[1]{}
    \newcommand{\YCinline}[1]{}
\else
    \newcommand{\YC}[1]{\todo[fancyline,color=Maroon!40]{YC: #1}\xspace}
    \newcommand{\YCinline}[1]{\textcolor{Maroon}{[YC: #1]}}
\fi

\DeclareMathOperator*{\argmin}{arg\,min}

\usepackage[normalem]{ulem}
\usepackage{soul}

\newcommand*\diff{\mathop{}\!\mathrm{d}}

\newcommand{\figref}[1]{Fig.~\ref{#1}}
\newcommand{\eqnref}[1]{\text{Eq.}~(\ref{#1})}
\newcommand{\secref}[1]{\S\ref{#1}}
\newcommand{\appref}[1]{Appendix~\ref{#1}}

\icmltitlerunning{Direct Acquisition Optimization for Low-Budget Active Learning}

\begin{document}

\twocolumn[
\icmltitle{
    Direct Acquisition Optimization for Low-Budget Active Learning
}



\icmlsetsymbol{equal}{*}

\begin{icmlauthorlist}
\icmlauthor{Zhuokai Zhao}{uchi-cs}
\icmlauthor{Yibo Jiang}{uchi-cs}
\icmlauthor{Yuxin Chen}{uchi-cs}
\end{icmlauthorlist}

\icmlaffiliation{uchi-cs}{Department of Computer Science, University of Chicago, Chicago IL, USA}

\icmlcorrespondingauthor{Zhuokai Zhao}{zhuokai@uchicago.edu}

\icmlkeywords{Machine Learning, ICML}

\vskip 0.3in
]



\printAffiliationsAndNotice{}  

\begin{abstract}
Active Learning (AL) has gained prominence in integrating data-intensive machine learning (ML)
models into domains with limited labeled data.
However, its effectiveness diminishes significantly when the labeling budget is low.
In this paper, we first empirically observe the performance degradation of existing AL 
algorithms in the low-budget settings, and then introduce \textit{Direct Acquisition 
Optimization} (\algname), a novel AL algorithm that optimizes sample selections based on 
expected true loss reduction.
%
Specifically, \algname utilizes influence functions to update model parameters and incorporates 
an additional acquisition strategy to mitigate bias in loss estimation. 
This approach facilitates a more accurate estimation of the overall error reduction, without 
extensive computations or reliance on labeled data. 
Experiments demonstrate \algname's effectiveness in low budget settings, outperforming 
state-of-the-arts approaches across seven benchmarks.
%
%
\end{abstract}

\section{Introduction}\label{sec:introduction}
Active learning (AL) explores how adaptive data collection can reduce the amount of 
data needed by machine learning (ML) models~\cite{settles2009active, ren2021survey}. 
It is particularly useful when labeled data is scarce or expensive to obtain, which 
significantly limits the adaptability of modern deep learning (DL) models due to their 
data-hungry nature~\cite{van2014modern}.
In these cases, AL algorithms selectively choose the most beneficial data points
for labeling, thereby maximizing the effectiveness of the training process even if the data
is limited in number.
In fact, AL has been broadly applied in many fields~\cite{adadi2021survey}, such as 
medical image analysis~\cite{budd2021survey}, astronomy~\cite{vskoda2020active}, and 
physics~\cite{ding2023active}, where unlabeled samples are plentiful but the process of labeling 
through human expert annotations or experiments is highly cost-intensive.
In these contexts, judiciously selecting samples for labeling can significantly lower 
the expenses involved in compiling the datasets~\cite{ren2021survey}. 

Many active learning algorithms have emerged over the past decades, with early 
seminal contributions from~\cite{lewis1995sequential, tong2001support, roy2001toward}, and a shift that focuses more on deep active learning - a branch of AL that targets more 
towards DL models in more recent years~\cite{huang2021deepal}. 
Depending on the optimization objective, AL algorithms can be classified into two categories. The first category includes heuristic objectives that are not exactly the same as the evaluation 
metric, i.e. error reduction.
Examples in this category are diversity~\cite{sener2017active}, 
uncertainty~\cite{gal2017deep}, and hybrids of both~\cite{ash2019deep}. 
Second category includes criteria that is exactly the same as the evaluation metric, 
where notable approaches include expected error reduction (EER)~\cite{roy2001toward} and 
its modern follow-up works~\cite{killamsetty2021glister, mussmann2022active}.
%

Despite the popularity of the first type of AL algorithms, we show in 
\secref{sec:motivations} that these methods often suffer heavily in low-budget settings, 
where the total (accumulative) sampling quota is less than 1\% of the number of unlabeled data points,
making them less suitable for the extreme data scarcity scenarios.
In terms of the methods from the second category, their higher running time and 
reliance on the availability of a \textit{validation} or \textit{hold-out} set remain 
significant limitations, constraining their applicability in many data-scarcity 
scenarios as well.
For example, EER~\cite{roy2001toward} re-trains the classifier for each candidate with all 
its possible labels, where in each time also evaluates the updated model on all the
unlabeled data, making its runtime intractable especially for deep neural networks.
And GLISTER~\cite{killamsetty2021glister}, despite being much more computationally 
efficient, requires a \textit{labeled, hold-out} set for its sample selection process, 
formulated as a mixed discrete-continuous bi-level optimization problem, to be optimized
properly.
%
%

While these constraints might not be a huge limitation a few years ago, it poses a more
important challenge currently as we are adopting deep learning models to more areas, 
where labeled data may be extremely expensive to acquire.
More importantly, it is also worth noticing that under these scenarios, the highly limited 
labeled data should have been better utilized for training than being reserved for AL 
algorithms.

Above limitations highlight a critical gap between the capabilities of current AL 
methodologies and the urgent demands from real-world applications, underscoring the need for 
developing novel AL strategies that can operate both relatively efficient while presenting 
little to none reliance on the labeled set.
To this end, we introduce \textbf{Direct Acquisition Optimization (\algname)}, a novel
AL algorithm that selects new samples for labeling by efficiently estimating the expected 
loss reduction.
Compared to EER and GLISTER, \algname solves the pain points of prohibitive running time and 
the reliance on a separate labeled set through utilizing \textit{influence 
function}~\cite{ling1984residuals} in model parameters updates, and a more accurate,
efficient unbiased estimator of loss reduction through importance-weighted sampling. 
%

To summarize, the contributions of this paper are: (1) an empirical analysis of 
existing AL algorithms under low budget settings; 
(2) a novel AL algorithm, Direct Acquisition Optimization (\algname), which optimizes sample 
selections based on expected error reduction while operating efficiently through influence 
function-based model parameters approximation and true overall reduced error estimation;
and (3) thorough experiments demonstrating \algname's superior performance in the
low-budget settings, out-performing current popular AL methods across seven benchmarks.
%
\section{Low-Budget Active Learning: A Motivating Case Study}\label{sec:motivations}
In this section, we provide an empirical analysis to demonstrate that commonly used 
heuristic-based AL algorithms do not work well under very low-budget settings.
Specifically, we analyze (1) uncertainty sampling methods including least
confidence~\cite{lewis1995sequential}, minimum margin~\cite{scheffer2001active}, 
maximum entropy~\cite{settles2009active}, 
and Bayesian Active Learning by Disagreement (BALD)~\cite{gal2017deep}; (2) diversity sampling 
methods such as Core-Set~\cite{sener2017active} and Variational Adversarial Active Learning 
(VAAL)~\cite{sinha2019variational}; and (3) hybrid method such as Batch Active learning by 
Diverse Gradient Embeddings (BADGE)~\cite{ash2019deep}.

\begin{figure}[t]
    \centering
    \includegraphics[width=\linewidth]{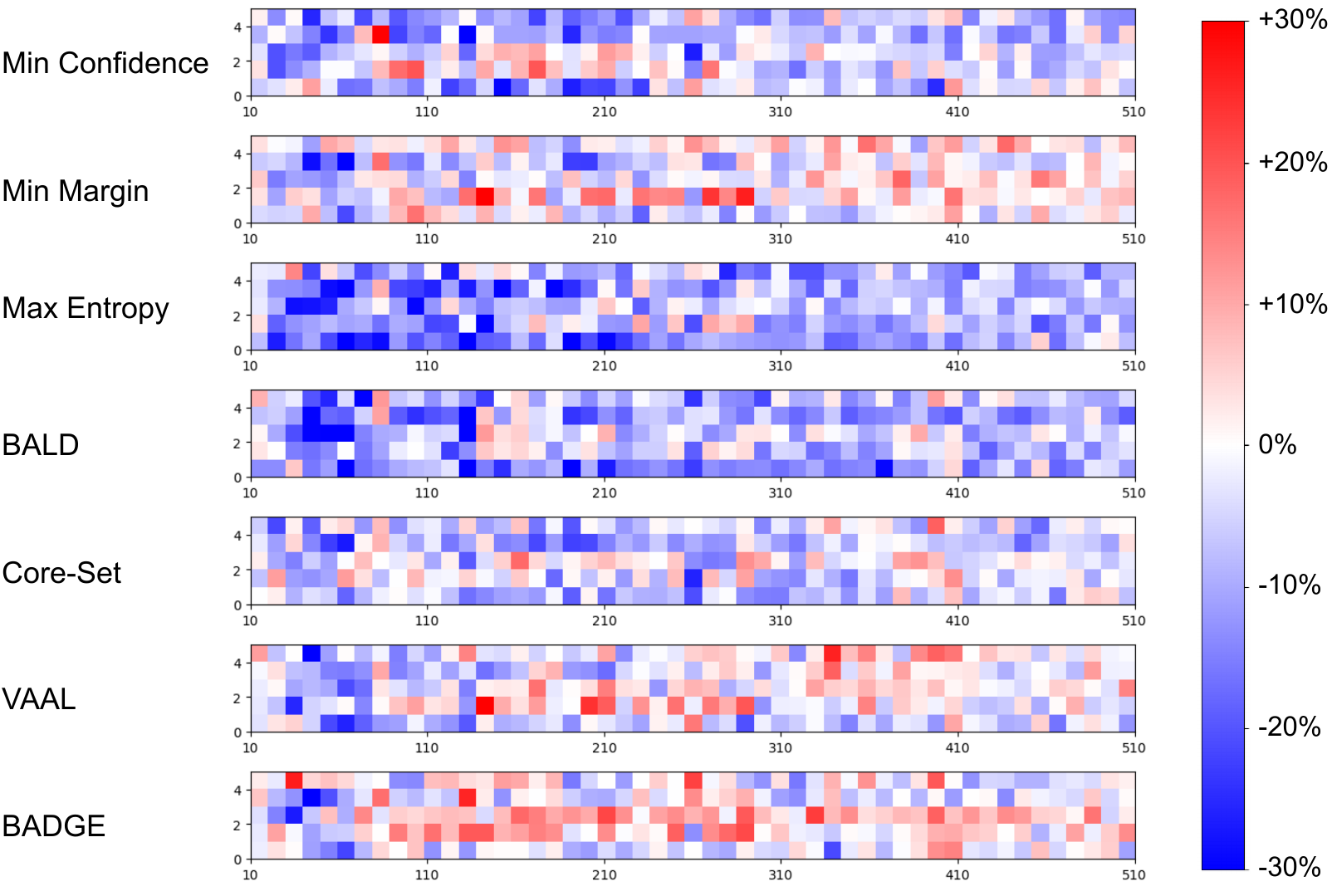}
    \caption{
        Existing methods fail to outperform random sampling with small budgets.
        This figure shows the relative performance between multiple methods and random acquisition.
        Within each subplot, $x$ axis represents the accumulative acquisition size, while
        $y$ axis indicates runs initiated with different random seeds.
        White color indicates on-par performance with random, blue indicates worse, and red
        indicates better.
    }
    \label{fig:motivation}
    \vspace{-0.2in}
\end{figure}

We test the above methods on the CIFAR10~\cite{krizhevsky2009learning} dataset starting with an 
initial labeled set with size $|\mathcal{L}_{\text{init}}| = 10$, and conducted 50 acquisition 
rounds where after each round $B = 10$ new samples are selected and labeled.
We use ResNet-18~\cite{he2016deep} as our training model across all methods.
And we repeated the acquisitions five times with different random seeds.
The results are visualized in \figref{fig:motivation}, where we plot the \textit{relative} 
performance between each method and random sampling acquisition through a diverging color map.

Aligning with the general perceptions that 
low-budget~\cite{mittal2019parting, hacohen2022active} and 
cold-start~\cite{zhu2019addressing, chandra2021initial} AL tasks are especially challenging, we
empirically observe that almost all popular AL algorithms fail to outperform the naive random
sampling when acquisition quota is less than 1\% (500 out of 50,000 in the case of CIFAR10) of 
the unlabeled size.
More specifically, when the quota is less than 0.2\% (less than 100 data points for CIFAR-10), all 
methods fail to reliably outperform random sampling (as the beginning of each heatmap in 
\figref{fig:motivation} are almost all blue), which greatly motivates the development of 
\algname.
We also include the more conventional line plot of the empirical analysis which may provide more 
detailed information of each run in \appref{subsec:lineplot}.
\section{Methodology}\label{sec:method}
Different from the heuristics-based AL algorithms that optimize criteria such as
diversity or uncertainty, \algname is built upon the EER formulation with the selection 
objective being the largest reduced error evaluated on the entire unlabeled set.
More specifically, \algname majorly improves upon two aspects: 
(1) instead of re-training the classifier, we employ influence
function~\cite{cook1982residuals}, a concept with rich history in statistical learning, to 
formulate the new candidate sample as a small perturbation to the existing labeled 
set, so that the model parameters can be estimated without re-training; 
and (2) instead of reserving a separate, relatively large labeled set for 
validation~\cite{killamsetty2021glister}, we sample a very small subset directly from the
\textit{unlabeled} set and estimate the loss reduction through bias correction.

\begin{figure}[t]
    \centering
    \includegraphics[width=\linewidth]{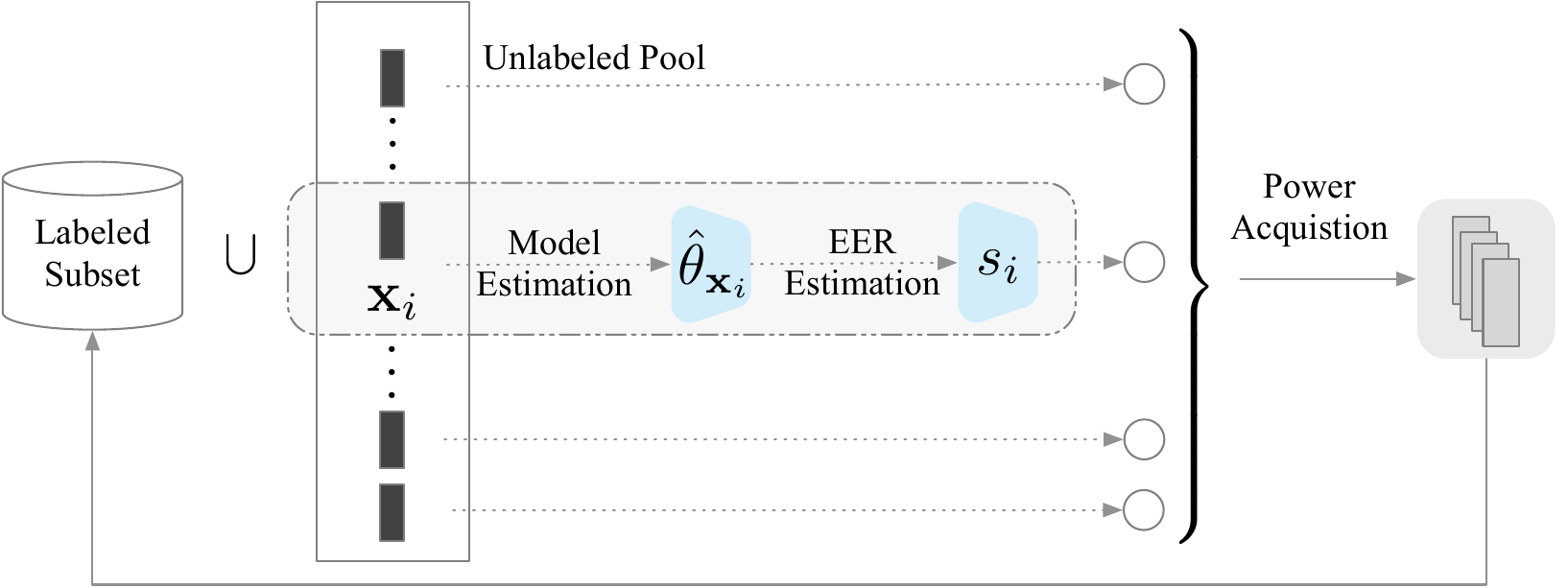}
    \caption{
        Schematic of the algorithmic framework of \algname.
    }
    \label{fig:illustration}
    \vspace{-0.15in}
\end{figure}

Essentially, when considering each candidate from the unlabeled set, we optimize the EER 
framework on two of its core components, which are model parameter update and true loss 
estimation.
Additionally, we upgrade EER, which only supports single sequential acquisition, to offer 
\algname in both single and batch acquisition variants by incorporating stochastic samplings to 
the sorted estimated loss reductions. We illustrate our algorithmic framework in 
\figref{fig:illustration}. 
In the following parts of this section, we first introduce a more formal problem statement 
in \secref{subsec:problem_statement}, and then dive into each specific component of \algname 
from \secref{subsec:label_estimate} to \secref{subsec:stochastic_sampling}.

\subsection{Problem Statement}\label{subsec:problem_statement}
The optimal sequential active learning acquisition function 
can be formulated as selecting a single $\mathbf{x}^{\text{train}}_t$ from the current unlabeled 
set $\mathcal{U}_t$ at each round $t$ such that 
%
\begin{equation}\label{eqn:formulation}
    \mathbf{x}^{\text{train}}_t 
    = \argmin_{\mathbf{x}_i \in \mathcal{U}_{t-1}} \mathbb{E}_{(y_i|f^*, \mathbf{x}_i)} 
    \left[
    L_{\text{true}}(f_{t|\mathbf{x}_i, y_i})
    \right]
\end{equation}
%
where $f^*$ represents an optimal oracle that maps from any unlabeled data 
$\mathbf{x}_i \in \mathcal{U}_{t-1}$ to its ground-truth label $y_i$, and 
$f_{t|\mathbf{x}_i, y_i}$
is the model that has been trained on the union of the current labeled set 
$\mathcal{L}_{t-1}$ and the current unlabeled candidate $\mathbf{x}_i \in \mathcal{U}_{t-1}$.
In addition, 
$L_{\text{true}}(f_{t|\mathbf{x}_i, y_i}) = \frac{1}{|\mathcal{U}_{t-1, i}|}\sum_{\mathbf{x} \in \mathcal{U}_{t-1, i}} \ell(\mathbf{x}; f_{t|\mathbf{x}_i, y_i})$
represents the loss estimator that can predict the \textit{unbiased} error of 
$f_{t|\mathbf{x}_i, y_i}$,
where $\ell$ denotes the loss of $f_{t|\mathbf{x}_i, y_i}$ evaluated on $\mathbf{x}_i$ (against its true label $f^*(\mathbf{x}_i)$).
It is numerically the same as if 
$f_{t|\mathbf{x}_i, y_i}$
has been tested on the entire unlabeled set 
$\mathcal{U}_{t-1, i}$, where $\mathcal{U}_{t-1, i}=\mathcal{U}_{t-1}\setminus\{\mathbf{x}_i\}$.
Such formulation represents the optimal AL algorithm and aligns with any existing 
sequential active learning algorithm ---
of which the goal is to select the new data point that 
can most significantly improve the current model performance~\cite{roy2001toward}.

Unfortunately, \eqnref{eqn:formulation} cannot be directly implemented in practice.
Because, first, we do not have access to the optimal oracle $f^*$ to reveal the label $y_i$ 
for each $\mathbf{x}_i \in \mathcal{U}_{t-1}$; 
second, even if we had $f^*$ and therefore $y_i$, we cannot afford the cost of retraining model 
$f_{t-1}$ on each $\mathcal{L}_{t-1} \cup \mathbf{x}_i$ to obtain the updated $f_{t|\mathbf{x}_i, y_i}$; 
and third, we do not have the unbiased true loss estimator $L_{\text{true}}$, which demands 
evaluating $f_{t|\mathbf{x}_i, y_i}$ on the entire $\mathcal{U}_{t-1, i}$.

Therefore, the goal of \algname is to solve the above challenges and efficiently and 
accurately approximate \eqnref{eqn:formulation} for the 
sample selection strategy.
It is also worth noting that, when $\mathbf{x}^{\text{train}}_t$ represents a \textit{set} of 
newly acquired data points, the above formulation becomes eligible for batch active learning, 
which is more suitable for deep neural networks~\cite{huang2021deepal}.

\subsection{Label Approximation via Surrogate}\label{subsec:label_estimate}
In this section, we address the first challenge when approximating \eqnref{eqn:formulation}.
As we do not know the true label or true label distribution $p(y | \mathbf{x}, f^*)$ 
of each unlabeled sample $\mathbf{x}$, the best we can do is provide an approximation for 
$p(y | \mathbf{x})$.
%
To this end, we introduce the concept of a \textit{surrogate}~\cite{kossen2021active}, 
which is a model parameterized by some potentially infinite set of parameters $\theta$. 
%
Specifically, $p(y | \mathbf{x})$ can be approximated using the marginal distribution 
$\pi(y | \mathbf{x}) = \mathbb{E}_{\pi(\theta)}[\pi(y | \mathbf{x}, \theta)]$ 
with some proposal distribution $\pi(\theta)$ over model parameters $\theta$. 
%
In other words, we have:
%
\begin{equation}\label{eqn:label_approximation}
    p(y|\mathbf{x}) \approx \int_\theta \pi(\theta)\pi(y|\mathbf{x},\theta) \diff \theta
\end{equation}
%
%
As the sample selection process continues, new labeled points should also be used to train and update
the surrogate model $\pi(\theta)$ for better approximation of the true outcomes. 
%
%

Although ideally, a more capable surrogate is preferred for better ground truth 
approximations, we acknowledge that the choice of surrogate model can be very sensitive to 
the computational constraints. 
Therefore, if running time is at center of the concerns during sample acquisitions, using 
$f_t$ at step $t$
also as the surrogate could be an efficient alternative, as we don't need to update a 
second model, nor do we need to run forward pass on the both models.
However, this will come with the cost that $\pi_t$ never disagrees with $f_t$, which causes 
performance degradation for the unbiased true loss estimation, which will be illustrated
with more details in \secref{subsec:loss_estimation}.
Therefore, in short, we do not recommend replicating $f_t$ as surrogate in practice, unless the 
computational constraint is substantial.

\subsection{Model Parameters Update without Re-training}\label{subsec:model_update}
At acquisition round $t$, suppose we have labeled set $\mathcal{L}_{t-1}$ and unlabeled 
set $\mathcal{U}_{t-1}$ as the results from the previous round $t-1$, and new sample 
$\mathbf{x}_i \in \mathcal{U}_{t-1}$ that is currently under consideration for 
acquisition, the goal of this section is to estimate the parameters of model $f_{t|\mathbf{x}_i, y_i}$ 
that could has been obtained after training $f_{t-1}$ on the combined dataset 
$\{\mathcal{L}_{t-1} \cup \mathbf{x}_i\}$.
In other words, if we suppose the conventional full training converges to parameters 
$\hat{\theta}_{\mathbf{x}_i}$, we have:
%
\begin{equation}\label{eqn:full_training}
    \hat\theta_{\mathbf{x}_i} = \arg\,\min_{\theta \in \Theta} \frac{1}{|\mathcal{L}_{t-1}|+1}\sum_{\mathbf{x} \in \{\mathcal{L}_{t-1} \cup \mathbf{x}_i\}} \ell(\mathbf{x}; \theta)
\end{equation}
where recall that $\ell(\mathbf{x}; \theta)$ denotes the loss of $\theta$ on $\mathbf{x}$.
The core of our approach is that, instead of re-training as showed in 
\eqnref{eqn:full_training}, we can approximate the effect of adding a new sample as 
upweighting the influence function by 
$\frac{1}{|\mathcal{L}_{t-1}|+1}$~\cite{koh2017understanding} and then directly estimate 
the updated model parameters.

Following~\cite{cook1982residuals},
we have the influence function defined as:
\begin{equation}
    \mathcal{I}_{\text{up}, \text{params}}(\mathbf{x}_i) \coloneqq \frac{d\hat{\theta}_{\epsilon, \mathbf{x}_i}}{d\epsilon}\bigg|_{\epsilon=0} = -H_{\hat{\theta}}^{-1} 
    \nabla_\theta \ell(\mathbf{x}_i; \hat{\theta})
\end{equation}
where $H_{\hat{\theta}}$ is the positive definite Hessian matrix~\cite{koh2017understanding}. 
%
%
Next, we can estimate the model parameters after adding this new sample $\mathbf{x}_i$, as:
\begin{equation}
\begin{split}
    \hat{\theta}_{\mathbf{x}_i} - \hat{\theta}
    &\approx \frac{1}{|\mathcal{L}_{t-1}|+1}\mathcal{I}_{\text{up}, \text{params}}(\mathbf{x}_i) 
    \\ 
    &= -\frac{1}{|\mathcal{L}_{t-1}|+1} H_{\hat{\theta}}^{-1} \nabla_\theta \ell(\mathbf{x}_i; \hat{\theta})
\end{split}
\end{equation}
where $\nabla_\theta \ell(\mathbf{x}_i; \hat{\theta})$ could be 
approximated as the expected gradient of sample $\mathbf{x}_i$: By a slight abuse of notation of the training loss function $\ell$, we denote
\begin{equation}\label{eqn:gradient_approximation}
    \nabla_\theta \ell(\mathbf{x}_i; \hat{\theta}) \approx \sum_{k=1}^K \nabla_\theta \ell(\mathbf{x}_i, \hat{y}_k; \hat{\theta}) \cdot \hat{p}_k
\end{equation}
In \eqnref{eqn:gradient_approximation}, $\hat{y}_k$ and $\hat{p}_k$ represent model's 
label prediction and likelihood (e.g. confidence) respectively while $K$ represents the 
total number of classes in the ground truths.

In practice, the inverse of $H_{\hat{\theta}}$ cannot be computed due to its prohibitive 
$O(np^2+p^3)$ runtime~\cite{liu2021influence}, with $p$ being the number of model 
parameters.
The computation unavoidably becomes especially intensive when $f$ is a deep neural network 
model~\cite{fu2018scalable}.
Luckily, we have two optimization methods, conjugate gradients (CG)~\cite{martens2010deep} 
and stochastic estimation~\cite{agarwal2017second} at our disposal.

\paragraph{Conjugate gradients.} 
As mentioned earlier, by assumption
we have $H_{\hat{\theta}} \succ 0$ and $\nabla_\theta \ell(\mathbf{x}'; \hat{\theta})$ as a vector.
Therefore, we can calculate the inverse Hessian vector product (IHVP) through first 
transforming the matrix inverse into an optimization problem, i.e.
\begin{equation}
    H_{\hat{\theta}}^{-1} \nabla_\theta \ell(\mathbf{x}_i; \hat{\theta}) \equiv \arg\,\min_t \; t^T H_{\hat{\theta}} t - v^Tt
\end{equation}
and then solving it with CG~\cite{martens2010deep}, which speeds up the runtime effectively
to $O(np)$.

\paragraph{Stochastic estimation.}
Besides CG, we can also efficiently compute the IHVP using the stochastic estimation 
algorithm developed by Agarwal et al.~\cite{agarwal2017second}.
From Neumann series, we have $A^{-1} \approx \sum_{i=0}^{\infty} (I-A)^i$ for any matrix $A$.
Similarly, suppose we define the first $j$ terms in the Taylor expansion of 
$H_{\hat{\theta}}^{-1}$ as 
\begin{equation}
    H_{\hat{\theta}, j}^{-1} = \sum_{i=0}^j (I-H_{\hat{\theta}})^i = I + (I - H_{\hat{\theta}})H_{\hat{\theta}, j-1}^{-1}
\end{equation}
we have
$H_{\hat{\theta}, j}^{-1} \rightarrow H_{\hat{\theta}}^{-1}$ as $j \rightarrow \infty$.
The core idea of the stochastic estimation is that the Hessian matrix $H_{\hat{\theta}}$ 
can be substituted with any unbiased estimation when computing $H_{\hat{\theta}}^{-1}$.
In practice, we sample $n_{\text{ihvp}}$ data points from the existing labeled set 
$\mathcal{L}_{t-1}$ and use $\nabla_{\theta}^2 \ell(\mathbf{x}_i; \hat{\theta})$ as the 
estimator of $H_{\hat{\theta}}$~\cite{liu2021influence}.
%
Notice that since $n_{\text{ihvp}}$ is usually very small (in our experiments we used 
$n_{\text{ihvp}}=8$), it does not create a constraint on the size of the current labeled set, 
which does not interfere with the low-budget settings.
%

Finally, we can approximate the model parameters after the addition of $\mathbf{x}_i$ as
\begin{equation}\label{eqn:parameter_update}
    \hat{\theta}_{\mathbf{x}_i} = \hat{\theta} - \frac{1}{n+1} H_{\hat{\theta}}^{-1} \nabla_\theta \ell(\mathbf{x}_i; \hat{\theta})
\end{equation}
which does not require any re-training.
And we will demonstrate in \secref{subsec:if_vs_sb} that this parameter update strategy 
provides much better approximations than the naive single backpropagation as seen 
in the existing AL literature~\cite{killamsetty2021glister}.

\subsection{Efficient Unbiased Loss Estimation}\label{subsec:loss_estimation}
Referring back to \eqnref{eqn:formulation}, the last challenge that we need to address 
is to gain access to the unbiased true loss estimator $L_{\text{true}}$.
In other words, we want to predict the \textit{true} performance of $f_{t|\mathbf{x}_i, y_i}$ on the 
unlabeled set $\mathcal{U}_{t, i}$ without exhaustive testing.
Strictly, such evaluation cannot be drawn until $f_{t|\mathbf{x}_i, y_i}$ is evaluated on the entire unlabeled 
set $\mathcal{U}_{t, i}$.
However, this is infeasible in practice.

Such approximation is typically carried out in other 
approaches~\cite{killamsetty2021glister, mussmann2022active} by randomly sampling a 
labeled validation set $\mathcal{V}$ at the beginning of the entire acquisition process, 
which will later be used for evaluations in all the subsequent acquisition episodes.
Despite the simplicity as well as being i.i.d., which makes the estimated loss unbiased 
by nature, this approximation method suffers from large variance as the size of 
$\mathcal{V}$ is usually much smaller than $\mathcal{U}$, which unavoidably hurts the
acquisition performance.
It is also contradictory to the goal of AL in general, especially under the low-budget
settings, as discussed in \secref{sec:introduction}.

Different from the existing works, we propose to sample a subset $\mathcal{C}$ from
current $\mathcal{L}_{t-1}$ in each acquisition round based on an alternative acquisition 
function, and then correct the bias in the loss induced from this acquisition function.
In the meantime, we also want to keep the variance low, so that the final corrected loss 
enjoys both low bias and low variance, which is more preferable than the zero bias but 
high variance that the random i.i.d. sampling has.

Specifically, continuing with the notations from \secref{subsec:problem_statement}, let 
$\mathcal{C} = \{\mathbf{x}_{t, 1}, \dots, \mathbf{x}_{t, m}, \dots, \mathbf{x}_{t, n_\mathcal{C}}\}$, where
$\mathcal{C} \subset \mathcal{U}_{t-1}$, be the subset containing 
$n_\mathcal{C}$ samples selected 
for this true loss estimation at each round $t$.~\citet{farquhar2021statistical} shows that if 
$\mathbf{x}_{t, m}$ is sampled in proportion to the true loss of each 
data point, the bias originated from this selection can be corrected through the Monte 
Carlo estimator 
$\hat{R}_{\text{LURE}}$\footnote{LURE stands for Levelled Unbiased Risk Estimator}.
%
Following our notations, it takes the form:
%
\begin{equation}\label{eqn:r_lure}
    \hat{R}_{\text{LURE}} = \frac{1}{n_{\mathcal{C}}}\sum_{m=1}^{n_{\mathcal{C}}}{v_m}
    \ell \left(\mathbf{x}_{t, m};f \right)
\end{equation}
where recall that $\ell$ denotes the loss of $f$, and the importance weight $v_m$ is
%
\begin{equation}\label{eqn:v_m}
    v_{m} = 1 + \frac{|\mathcal{U}_{t-1}|-n_{\mathcal{C}}}{|\mathcal{U}_{t-1}|-m}\left(\frac{1}{(|\mathcal{U}_{t-1}|-m+1)q^*_t(m)} - 1\right)
\end{equation}
with $q^*_t(m)$ being the acquisition distribution of index $m$ at round t.
Importantly, the variance can be significantly reduced if the acquisition 
distribution $q^*_t(m)$ is proportion to the true loss of each data point.
Again, this is not feasible as we do not have access to the labels for $\mathcal{U}_{t-1}$.
However, following~\citet{kossen2021active}, we can approximate $q^*_t(m)$ with 
\begin{equation}
    q_t(m) = -\sum_y \pi(y|\mathbf{x}_{t, m})\log f(\mathbf{x}_{t, m})
\end{equation}
for classification tasks when the loss function is the cross-entropy loss, and where $\pi$ is 
conveniently just our surrogate discussed in \secref{subsec:label_estimate}.
Referring back to the discussion we had on choosing a good surrogate $\pi$,
with $f(\mathbf{x})$ being designed to approximate $p(y|\mathbf{x})$
as well, the surrogate $\pi$ should ideally be different from $f$ so that more diversity is 
introduced in the acquisitions.

To put all the components together, our loss correction process involves selecting samples 
in $\mathcal{C}$ based on 
%
\begin{equation}\label{eqn:correction_sampling}
    \mathbf{x}_{t,m} \propto -\sum_y \pi_{t-1}(y|\mathbf{x}) \log f_{t-1}(y|\mathbf{x})
\end{equation}
where $\pi_{t-1}$ is the surrogate model $\pi$ at round $t-1$. 
%
Finally, the corrected loss $s_i$ can be approximated using $\hat{R}_{\text{LURE}}$ as 
\begin{equation}\label{eqn:s_i}
    s_i = \frac{1}{n_{\mathcal{C}}}\sum_{m=1}^{n_{\mathcal{C}}} \hat{v}_m 
       \ell \left(\mathbf{x}_{t, m};f_{t} \right)
\end{equation}
where $\hat{v}_m$, which depends on the choice of $\mathbf{x}_{t, m}$, is
the approximated version of the original $v_m$ defined in \eqnref{eqn:v_m}.
Specifically, $\hat{v}_m$ takes the form
\begin{equation}
    \hat{v}_{m} = 1 + \frac{|\mathcal{U}_{t-1}|-n_{\mathcal{C}}}{|\mathcal{U}_{t-1}|-m}\left(\frac{1}{(|\mathcal{U}_{t-1}|-m+1)q_t(m)} - 1\right)
\end{equation}
where $q_t(m)$ is the acquisition function defined in \eqnref{eqn:correction_sampling}.

\subsection{Batch Acquisition via Stochastic Sampling}\label{subsec:stochastic_sampling}
In \secref{subsec:problem_statement}, we briefly discussed that when 
$\mathbf{x}^{\text{train}}_t$ represents a set of data points (instead of a single one), the 
formulation in \eqnref{eqn:formulation} essentially represents the \textit{batch} active 
learning scenario.
Suppose the acquisition budget per round is $k$, although selecting the top $k$ samples with the 
lowest estimated losses (or highest expected error reduction) is straightforward, this approach 
is sub-optimal. 
This is because top-$k$ acquisition, while effective to some degree due to its greedy nature, 
overlooks the crucial interactions among data points in batch acquisitions. 
Specifically, while aiming to select the most informative unlabeled points, top-$k$ acquisition 
may lead to redundant choices, diminishing the overall benefit of the acquisition.
%
%
%

Inspired by~\citet{kirsch2021stochastic}, we propose to similarly perturb the original 
ranking of the estimated true losses so that the batch sampling provides better 
acquisitions when the most informative data points may be duplicated.
Suppose at acquisition episode $t$, we rank the set of estimated true loss of each unlabeled 
data point in ascending orders as 
$\{\hat{l}_{\text{true}, i}\}_{\mathbf{x}_i \in \mathcal{U}_{t-1}}$, 
such that $\hat{l}_{\text{true}, i} \leq \hat{l}_{\text{true}, j}, \forall i \leq j \text{ and } 
\mathbf{x}_i, \mathbf{x}_j \in \mathcal{U}_{t-1}$, we can perturb the ranking with three 
strategies: soft-rank, soft-max, and power acquisition, to improve batch performance from the 
naive top-$k$ sampling.

\paragraph{Soft-rank acquisition.}\hspace{2pt}
Soft-rank acquisition relies on the relative ordering of the scores while ignoring the 
absolute score values.
It samples the data point ranked at index $i$ with probability 
$p_{\text{softrank}}(i) = i^{-\beta}$, where $\beta$ is the ``coldness" parameter and is 
kept as 1 throughout this paper.
It is not hard to notice that $p_{\text{softrank}}(i)$ is invariant to 
$\hat{l}_{\text{true}, i}$, as long as the relative ranking remains the same.
More conveniently, with sampled Gumbel noise $\epsilon_i \sim \text{Gumbel}(0; \beta^{-1})$, 
taking the top-$k$ data points from the perturbed ranked list 
\begin{equation}
    \hat{l}^{\text{softrank}}_{\text{true}, i} = -\log i + \epsilon_i
\end{equation}
is equivalent to sampling $p_{\text{softrank}}(i)$ 
without replacement~\cite{huijben2022review}.

\paragraph{Soft-max acquisition.}\hspace{2pt}
In contrast to soft-rank, soft-max acquisition uses the actual scores, i.e., the estimated
true losses, instead of their relative orderings.
However, this acquisition does not rely on the semantics of the actual values, resulting 
in the transformed true loss simply being:
\begin{equation}
    \hat{l}^{\text{softmax}}_{\text{true}, i} = \hat{l}_{\text{true}, i} + \epsilon_i
\end{equation}
where $\epsilon_i$ remains the same Gumbel noise as in the soft-rank acquisition.
Statistically, choosing the top-$k$ data points from this perturbed ranked list is equivalent
to sample from $p_{\text{softmax}}(i) = e^{\beta i}$ without replacement.

\paragraph{Power acquisition.}\hspace{2pt}
While neither soft-rank or soft-max acquisitions take the semantic meaning of the actual
score values into account when designing the acquisition distribution, power acquisition
uses the value directly when determining the perturbed values.
Specifically, the power acquisition perturbs the scores as 
\begin{equation}
    \hat{l}^{\text{power}}_{\text{true}, i} = \log \hat{l}_{\text{true}, i} + \epsilon_i
\end{equation}
where again $\epsilon_i$ is the Gumbel noise, and choosing the top-$k$ indices from this 
new list is equivalent to sampling from $p_{\text{power}}(i) = i^{\beta}$ without replacement.
Results comparing \algname with different batch acquisition strategies discussed above are 
showed in \appref{subsec:appendix_batch}.
%

Combining all the components, 
the pseudocode of \algname is summarized in Algorithm~\ref{alg:dao}.
\setlength{\textfloatsep}{12pt}
\begin{algorithm}[t]
    \caption{Direct Acquisition Optimization (\algname)}
    \label{alg:dao}
    \begin{algorithmic}[1]
       \INPUT Episode $t$, unlabeled set $\mathcal{U}_{t-1}$, 
       labeled set $\mathcal{L}_{t-1}$, model $f_{t-1}$, surrogate $\pi_{t-1}$,  
       budget $k$, $n_{\text{ihvp}}$ (\secref{subsec:model_update}), 
       and $n_{\mathcal{C}}$ (\secref{subsec:loss_estimation})
       \OUTPUT Acquisition set 
       $\mathcal{A}_t = \{\mathbf{x}^{\text{train}}_{t, 1}, \dots, \mathbf{x}^{\text{train}}_{t, k}\}$
       \COMMENT{\eqnref{eqn:formulation}}
       \STATE Approximate $p(y|\mathbf{x})$ for all $\mathbf{x} \in \mathcal{U}_{t-1}$ 
       \COMMENT{\secref{subsec:label_estimate}, \eqnref{eqn:label_approximation}}
       \STATE Initialize array $S$ where $|S| = |\mathcal{U}_{t-1}|$
       \FOR{$i=1$ {\bfseries to} $|\mathcal{U}_{t-1}|$}
       \STATE Let $\mathcal{U}_{t, i} = \mathcal{U}_{t-1} \setminus \{\mathbf{x}_i\}$
       \STATE Randomly sample $n_{\text{ihvp}}$ data points from $\mathcal{U}_{t, i}$
       \STATE Approximate parameters of $f_{t|\mathbf{x}_i, y_i}$
       \COMMENT{\secref{subsec:model_update}, \eqnref{eqn:parameter_update}}
       \STATE Acquire $n_c$ samples from $\mathcal{U}_{t, i}$
       \COMMENT{\secref{subsec:loss_estimation}, \eqnref{eqn:correction_sampling}}
       \STATE Compute $s_i$ and add to $S$
       \COMMENT{\secref{subsec:loss_estimation}, \eqnref{eqn:s_i}}
       \ENDFOR
       \STATE Sort $S$ in ascending order
       \IF{$k > 1$}
       \STATE Perturb $S$ 
       \COMMENT{Methods showed in \secref{subsec:stochastic_sampling}}
       \ENDIF
       \STATE Return top-$k$ samples in $S$ as $\mathcal{A}_t$
    \end{algorithmic}
\end{algorithm}
\section{Experiments}\label{sec:experiments}
We evaluate \algname on seven classification benchmarks including 
digit recognition datasets MNIST~\cite{lecun1998gradient},
Street-View House Numbers recognition (SVHN)~\cite{sermanet2012convolutional},
object classification datasets STL-10~\cite{coates2011analysis}, CIFAR-10, 
CIFAR-100~\cite{krizhevsky2009learning},
as well as domain-specific datasets Fashion-MNIST~\cite{xiao2017fashion} 
and Stanford Cars (Cars196)~\cite{krause20133d}.
%
%


\subsection{Experimental Setup}\label{subsec:experiment_setup}
\paragraph{Baselines.}\hspace{2pt} 
To ensure fair comparisons, besides baseline methods that we empirically surveyed
in \secref{sec:motivations}, we also include other state-of-the-arts AL methods, including
Deep Bayesian Active Learning (DBAL)~\cite{gal2017deep} 
and GLISTER~\cite{killamsetty2021glister}, where GLISTER is a direct competitor that
also optimizes the EER framework.
%
%

For all the baselines, we used the default/recommended parameters and their official 
implementations if publically available.
In terms of earlier works such as least confidence~\cite{lewis1995sequential}, 
minimum margin~\cite{scheffer2001active}, and maximum entropy~\cite{settles2009active},
we used the peer-reviewed deep active learning framework DeepAL+~\cite{zhan2022comparative}.
All experiments are repeated ten times with different random seeds.

\paragraph{Implementation details.}\hspace{2pt}
Throughout the experiment section, we set ResNet-18~\cite{he2016deep} as the model 
$f$ to be trained from scratch.
We employed VGG16~\cite{simonyan2014very}, initialized with random weights, as our 
surrogate $\pi$.
We used stochastic estimation~\cite{agarwal2017second}
when estimating the updated model parameters, as discussed in \secref{subsec:model_update}.
%
We choose $n_{\text{ihvp}} = 8$ when approximating 
the unbiased estimator of $H_{\hat{\theta}}$, 
and set $n_{\mathcal{C}} = 16$ for biased 
loss correction as in 
\secref{subsec:loss_estimation}.

\subsection{Digit Recognition}\label{subsec:digit_recognition}
First, we demonstrate \algname's effectiveness through two digit recognition benchmarks: MNIST~\cite{lecun1998gradient} and SVHM~\cite{sermanet2012convolutional}.
MNIST is a collection of handwritten digits consisting of 60k training and 10k test 
images, while SVHN is a more challenging dataset containing over 600k real-world house 
numbers images taken from street views.
Both datasets contain 10 classes corresponding to digits from 0 to 9.

\begin{figure*}[t]
    \centering
    \includegraphics[width=0.6\textwidth]{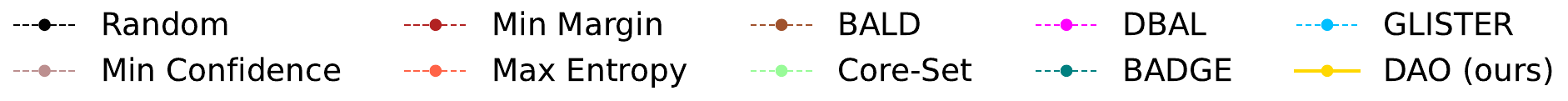}\\
    \begin{subfigure}[h]{0.24\textwidth}
        \centering
        \includegraphics[width=\textwidth]{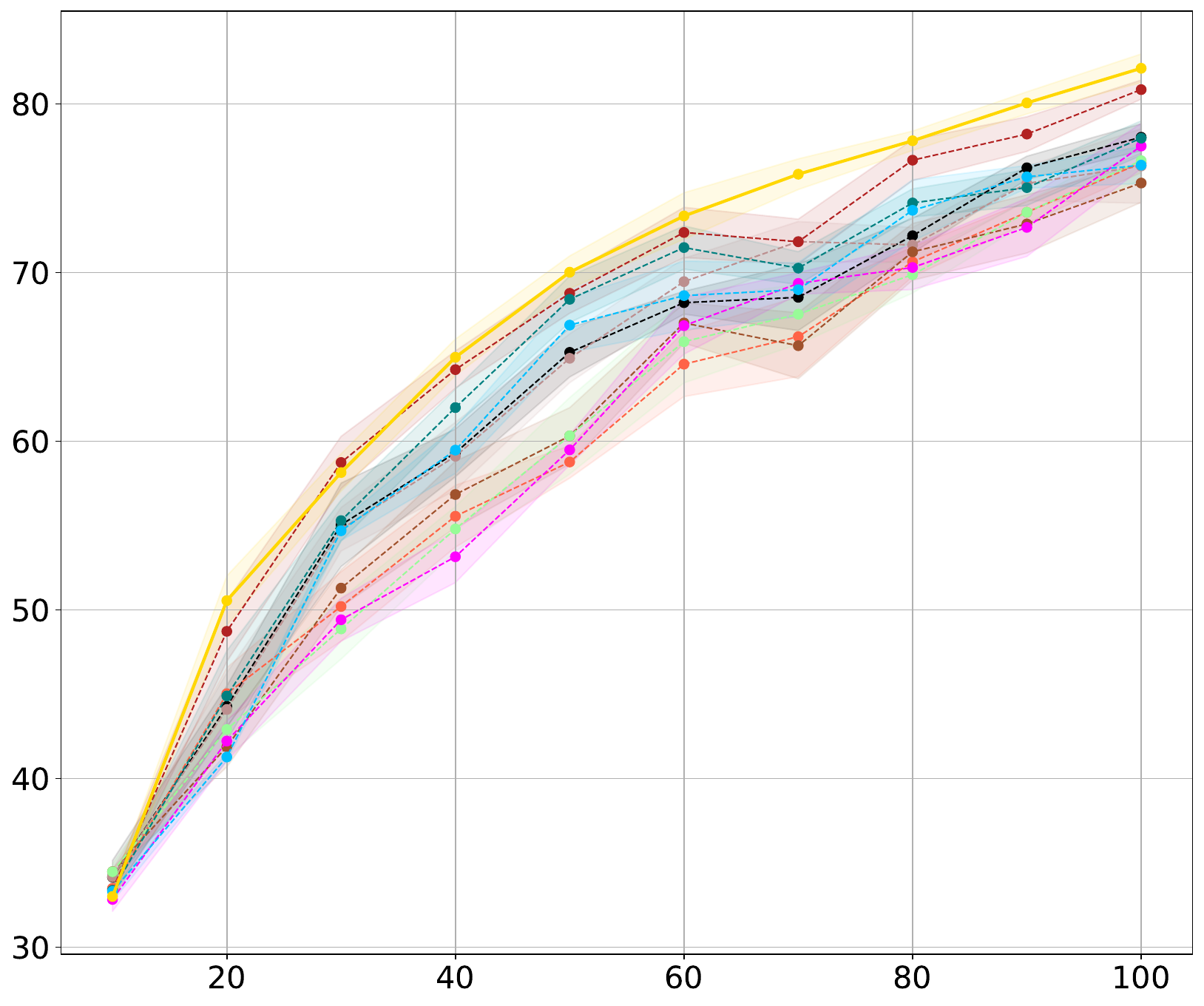}
        \caption{MNIST}
        \label{subfig:mnist_result}
    \end{subfigure}
    \begin{subfigure}[h]{0.24\textwidth}
        \centering
        \includegraphics[width=\textwidth]{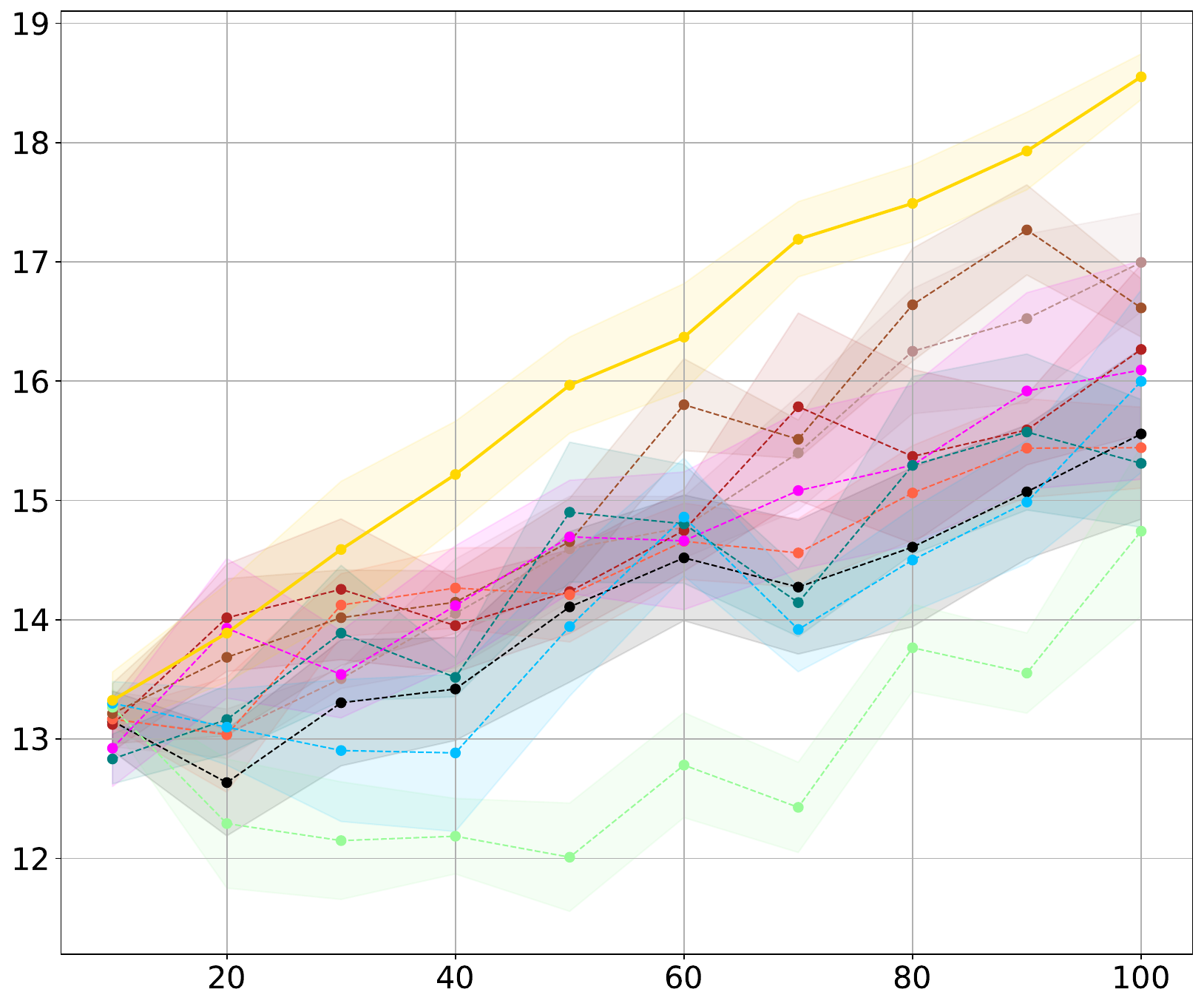}
        \caption{SVHN ($B = 10$)}
        \label{subfig:svhn_10_result}
    \end{subfigure}
    \begin{subfigure}[h]{0.24\textwidth}
        \centering
        \includegraphics[width=\textwidth]{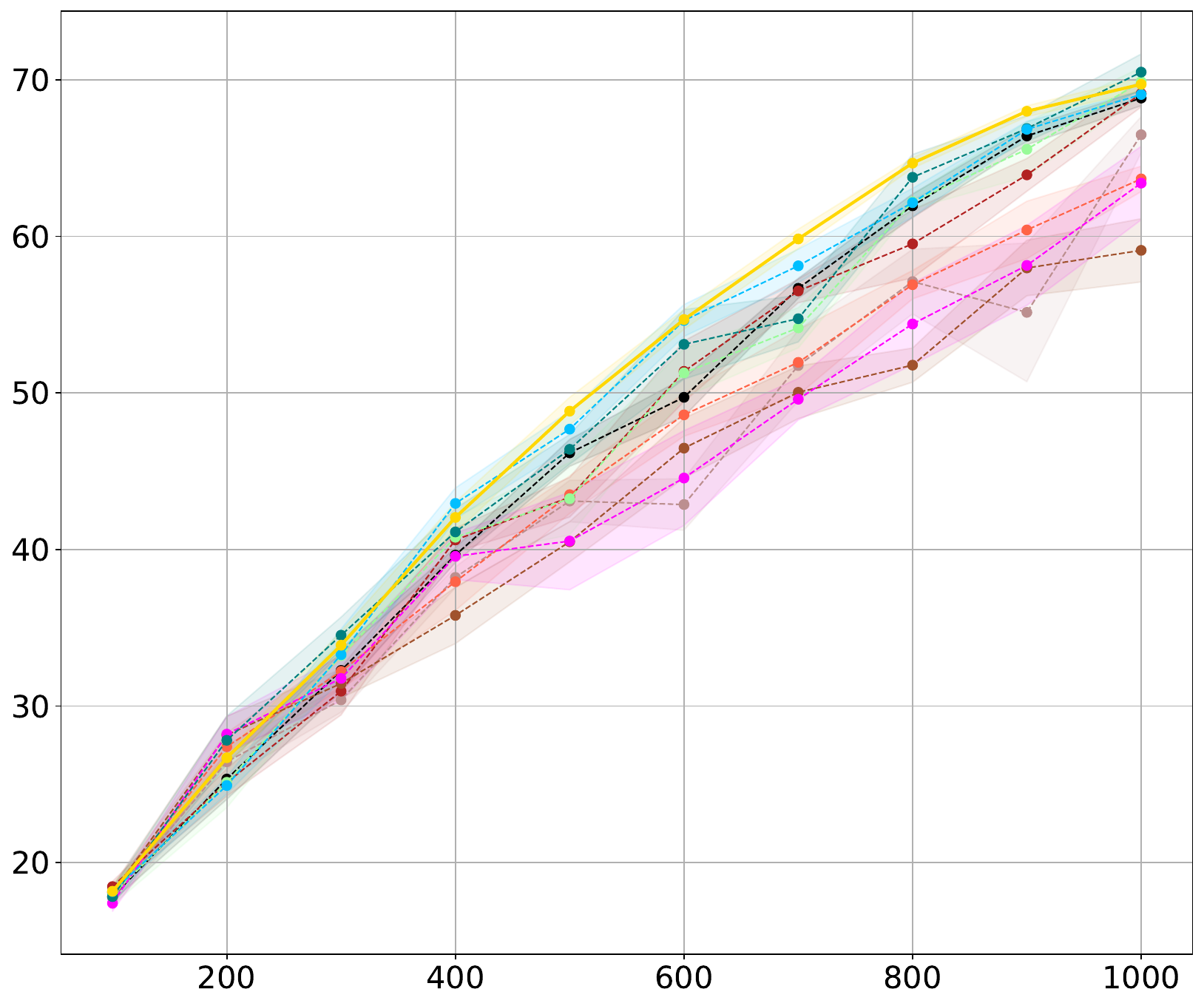}
        \caption{SVHN ($B = 100$)}
        \label{subfig:svhn_100_result}
    \end{subfigure}
    \begin{subfigure}[h]{0.245\textwidth}
        \centering
        \includegraphics[width=\textwidth]{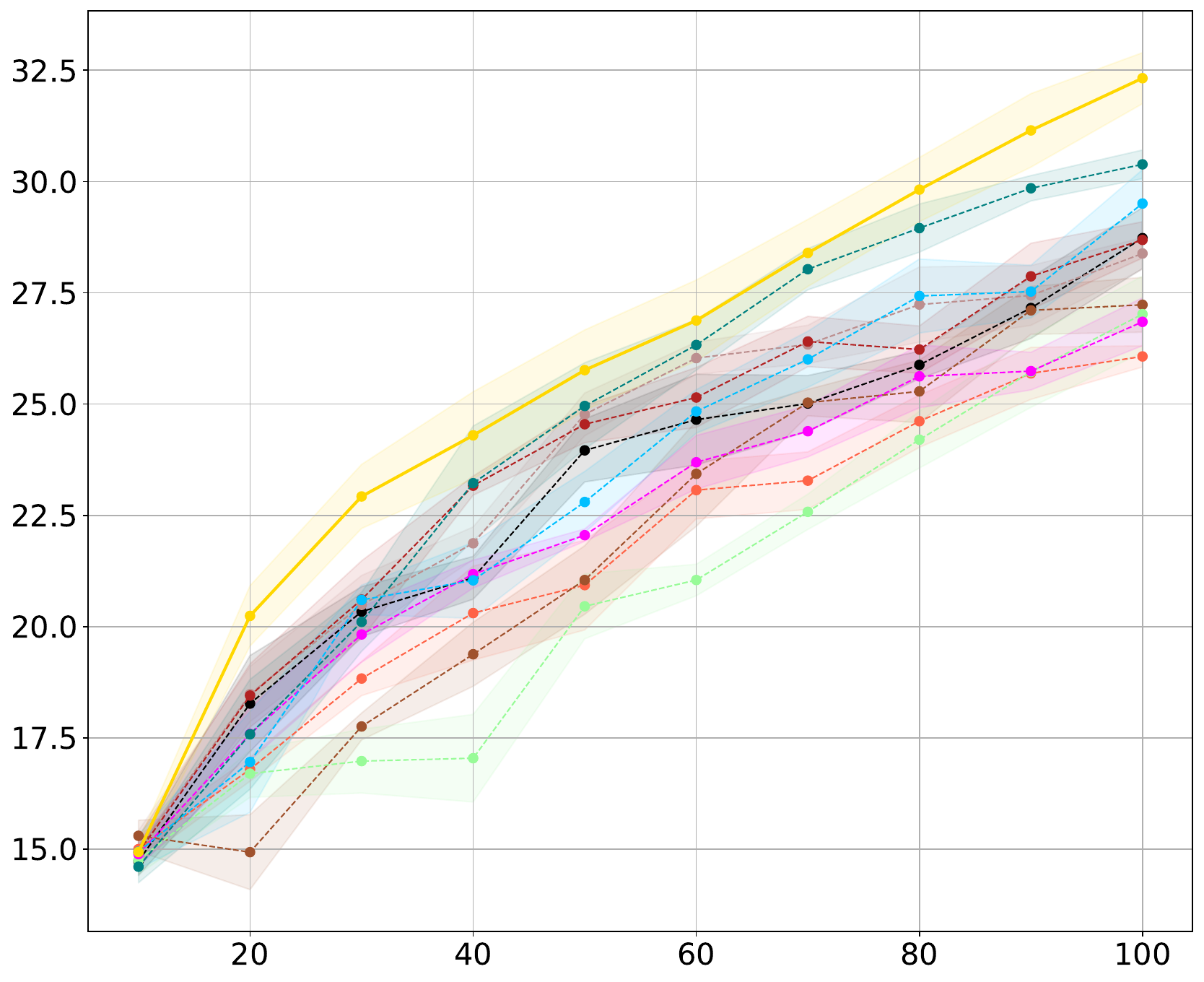}
        \caption{STL-10}
        \label{subfig:stl_10_result}
    \end{subfigure}
    \begin{subfigure}[h]{0.24\textwidth}
        \centering
        \includegraphics[width=\textwidth]{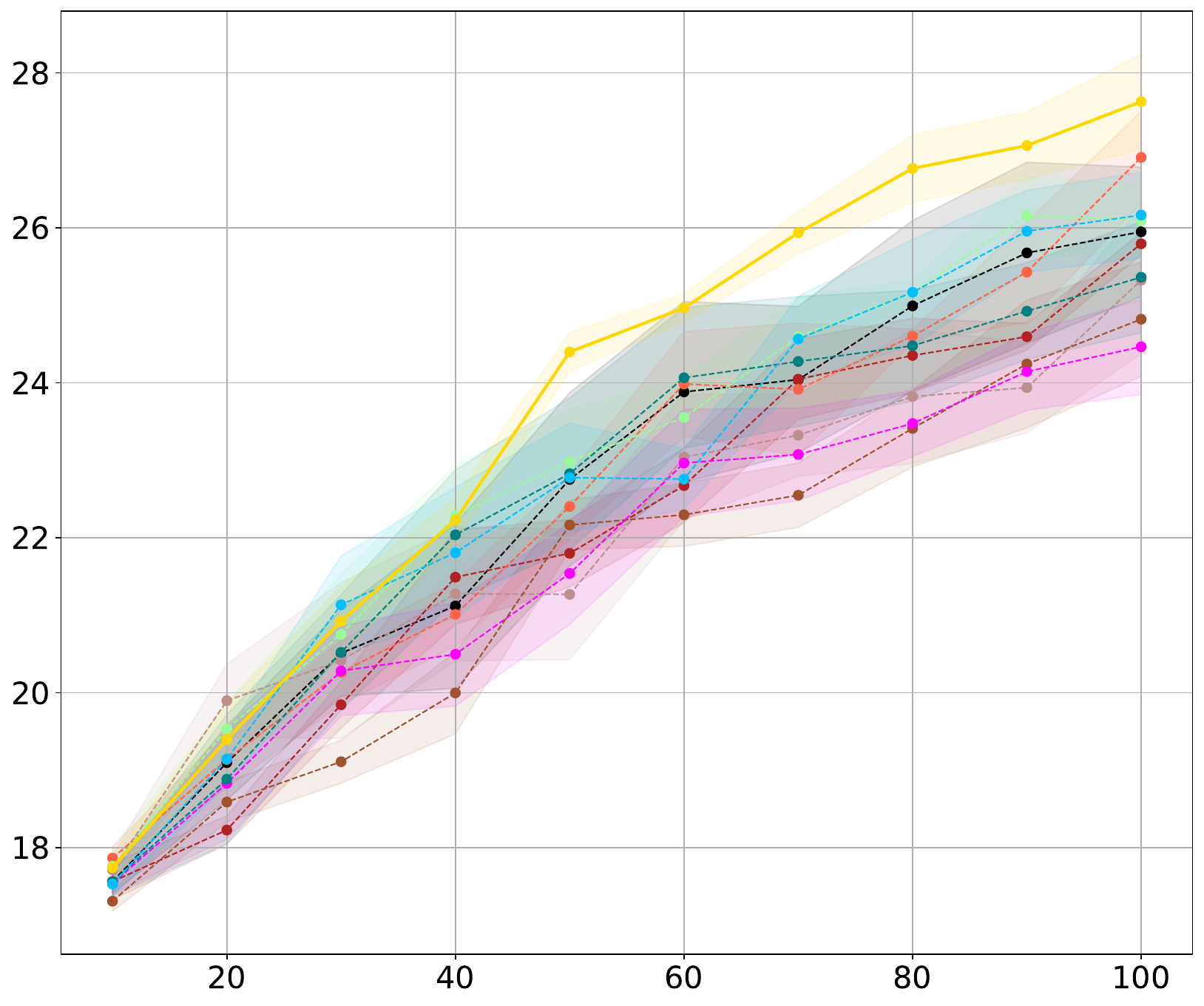}
        \caption{CIFAR-10}
        \label{subfig:cifar_10_result}
    \end{subfigure}
    \begin{subfigure}[h]{0.24\textwidth}
        \centering
        \includegraphics[width=\textwidth]{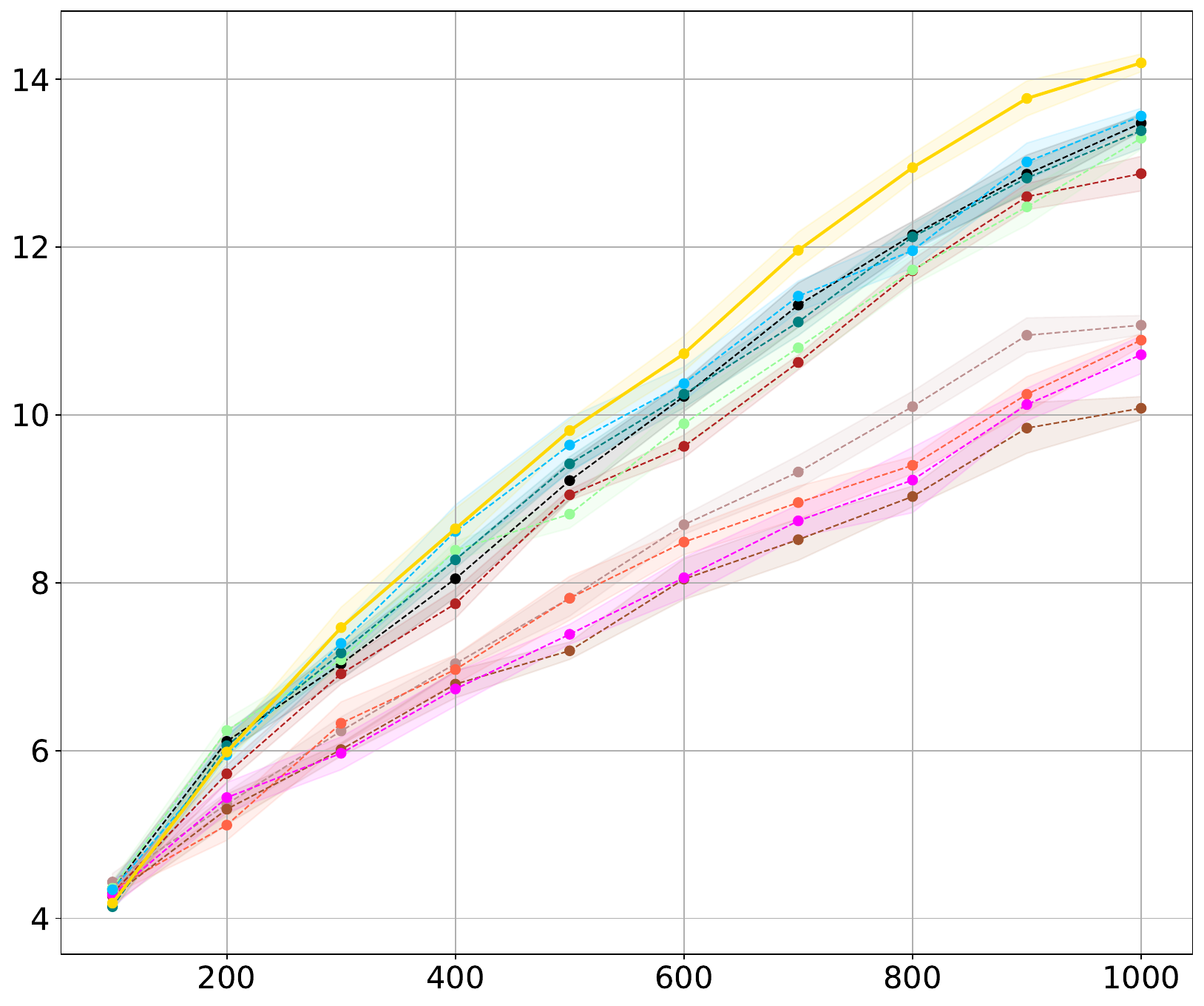}
        \caption{CIFAR-100}
        \label{subfig:cifar_100_result}
    \end{subfigure}
    \begin{subfigure}[h]{0.24\textwidth}
        \centering
        \includegraphics[width=\textwidth]{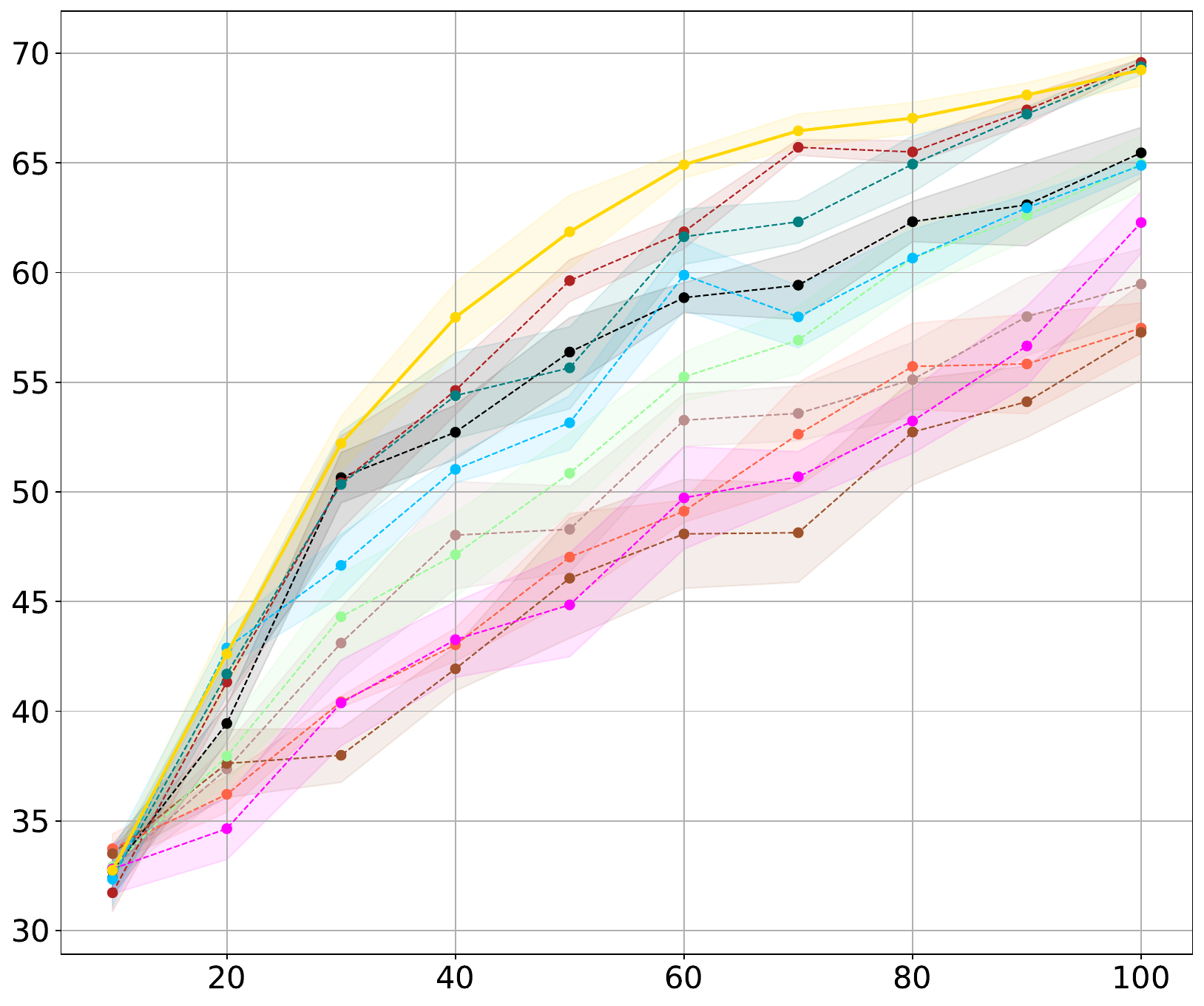}
        \caption{FashionMNIST}
        \label{subfig:fashion_mnist_result}
    \end{subfigure}
    \begin{subfigure}[h]{0.243\textwidth}
        \centering
        \includegraphics[width=\textwidth]{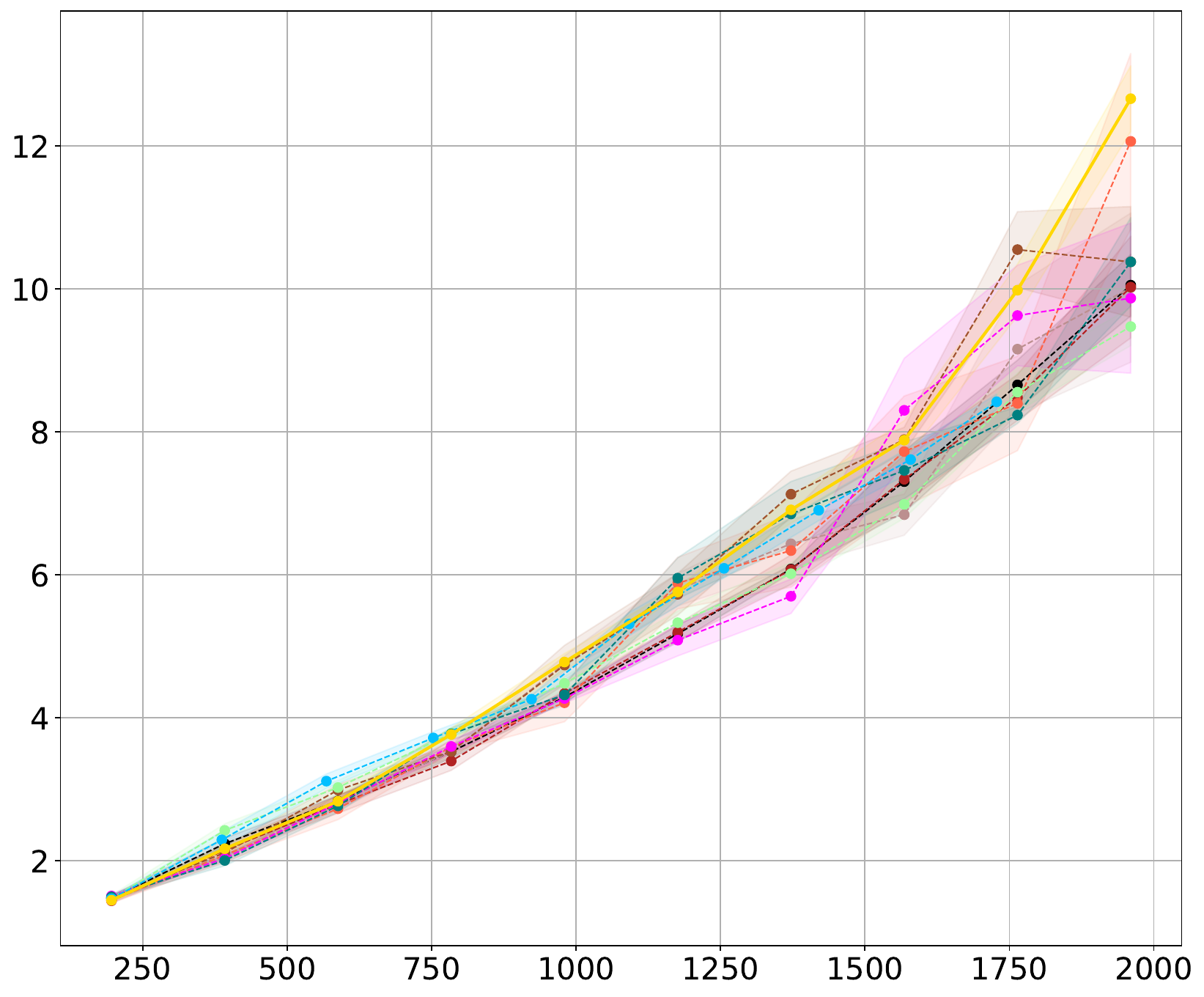}
        \caption{StanfordCars}
        \label{subfig:stanford_cars_result}
    \end{subfigure}
    \caption{
        Experiment results comparing \algname with existing AL algorithms across seven
        benchmarks.
        In all subplots, horizontal axis represents the accumulative size of the labeled set, 
        while vertical axis indicates classification accuracy.
    }
    \label{fig:all_results}
    \vspace{-0.15in}
\end{figure*}

Based on the insights from \secref{sec:motivations}, we define a general rule of 
low-budget setting as \textit{one image per class}, which translates to initial label size 
$|\mathcal{L}^{\text{MNIST}}_{\text{init}}| = 10$ and per-episode budget
$B_{\text{MNIST}} = 10$ for MNIST.
Given that SVHN is more challenging, and there are ten times more unlabeled images than
in MNIST (600k vs. 60k), we experiment both 
$|\mathcal{L}^{\text{SVHN}}_{\text{init}}| = 10, B_{\text{SVHN}} = 10$
and
$|\mathcal{L}^{\text{SVHN}}_{\text{init}}| = 100, B_{\text{SVHN}} = 100$ for SVHN.
The results are showed in \figref{subfig:svhn_10_result} and~\ref{subfig:svhn_100_result}.

\subsection{Object Classification}\label{subsec:object_classification}
Next, we assess \algname on more general and complex object classification tasks.
STL-10~\cite{coates2011analysis} is a benchmark dataset derived from labeled examples
in the ImageNet~\cite{deng2009imagenet}.
Specifically, STL-10 contains 5k labeled 96$\times$96 color images spread across 10 classes,
as well as 8k images in the test split.
%
%
CIFAR-10~\cite{krizhevsky2009learning} contains a collection of 60k 32x32 color 
images in 10 different classes, with 6k images per class.
CIFAR-100 is similar to CIFAR-10, but covers a much wider range, containing 100 classes 
where each class holds 600 images.

Continuing with the low-budget setting (\textit{1 image per class}), we have
$|\mathcal{L}^{\text{STL-10}}_{\text{init}}| = 10$, $B_{\text{STL-10}} = 10$
for STL-10,
$|\mathcal{L}^{\text{CIFAR-10}}_{\text{init}}| = 10$, $B_{\text{CIFAR-10}} = 10$
for CIFAR-10
and $|\mathcal{L}^{\text{CIFAR-100}}_{\text{init}}| = 100$, $B_{\text{CIFAR-100}} = 100$
for CIFAR-100.
The results are showed in \figref{subfig:stl_10_result},~\ref{subfig:cifar_10_result} 
and~\ref{subfig:cifar_100_result} respectively.

\subsection{Domain Specific Tasks}\label{subsec:domain_specific}
The last part of our experiments involves case studies on applying \algname to 
domain-specific tasks, which simulates many real-world applications.
Specifically, we use FashionMNIST~\cite{xiao2017fashion} and 
StanfordCars~\cite{krause20133d}, also known as Cars196, in this experiment.
FashionMNIST is structure-wise similar to MNIST, comprising 28$\times$28 images of 
70k fashion products from 10 categories, with 7k images per category.
The training set contains 60k images, while the test set includes the rest.
StanfordCars is a large collection of car images, containing 16,185 images with a 
near-balanced ratio on the train/test split, resulting in 8,144 and 8,041 images 
for training and testing.
There are 196 classes in total, where each class consists of the year, make, model of 
a car (e.g., 2012 Tesla Model S).
The results of both datasets are showed in \figref{subfig:fashion_mnist_result} 
and~\ref{subfig:stanford_cars_result}.

\subsection{Discussion}\label{subsec:discussion}
\looseness -1 
From \figref{fig:all_results}, we notice that the proposed \algname outperforms
popular AL state-of-the-arts by a clear margin across all seven benchmarks.
Especially, with SVHN, when the budget is extremely low ($B=10$, which is 
0.0017\% of the unlabeled size), \algname leads the performance by a very large
gap, indicating its superior capability in the low-budget setting.
Such performance does not degrade much as the budget constraint is relaxed. 
As shown in \figref{subfig:svhn_100_result}, \algname still performs relatively well.
The only experiment that \algname does not improve as much 
is the StanfordCars.
However, the accuracy improvement from \algname is more smooth and has less variance,
indicating better robustness when applied to the more challenging (StanfordCars has 196 classes) 
applications.
%

%

\section{Component Analysis and Ablation Studies}\label{sec:component_analysis}
We now analyze specific components of \algname and conduct ablation studies
on the strategies proposed for model parameters approximation (\secref{subsec:if_vs_sb}) 
and true loss prediction (\secref{subsec:bias_vs_random}).
%

\subsection{Accuracy on Model Approximation}\label{subsec:if_vs_sb}
First, we assess if estimating the model parameters updates through modelling the effect of 
adding a new sample as upweighting the influence function provides a more accurate model 
performance approximation than using single backpropagation as seen in the existing
work~\cite{killamsetty2021glister}.
Specifically, we conduct the experiments on CIFAR-10~\cite{krizhevsky2009learning}, 
with initial labeled size $|\mathcal{L}^{\text{CIFAR-10}}_{\text{init}}| = 100$ 
(randomly sampled from the train split), per-episode budget $B_{\text{CIFAR-10}}=1$, 
and number of acquisition episode $E=25$.
We compare the updated models performance (accuracy) on the test split of CIFAR-10.
Different from the experiments in \secref{sec:experiments}, 
we do not apply any AL algorithm when acquiring the sample in each round.
Instead, we randomly choose $B$ sample in each acquisition round from the 
unlabeled set and then update the models through both methods with the same selected sample.

To access the difference between models updated with our influence function-based method 
and single backpropagation, we compute the mean squared error (MSE) between the performance 
of each model and the model updated by conventional full training, which is defined in 
\eqnref{eqn:full_training}.

Based on the result showed in \figref{subfig:model_update_mse}, we see that the 
proposed method provides more accurate (smaller mean and median) and more robust (smaller std.) 
model approximations than single backpropagation, contributing to the performance 
gain we observe in \secref{sec:experiments}.

\subsection{Bias Correction vs. Random Sampling}\label{subsec:bias_vs_random}
Next, we conduct ablation studies on replacing the proposed loss estimation 
(\secref{subsec:loss_estimation}) with the average loss of randomly sampled data points.
More specifically, we replace the estimated loss $s_i$ from averaging the corrected
loss (\eqnref{eqn:r_lure}) of the acquired samples via an alternative acquisition criteria
(\eqnref{eqn:correction_sampling}) with averaging losses of the samples acquired uniformly, 
i.e., at round $t$, we have
%
    $s^{\text{random}}_i = \frac{1}{M_{\text{random}}} \sum_{m=1}^{M_{\text{random}}} 
    \ell(\mathbf{x}_{t,m}; f_t)$
%
where $\textbf{x}_{t, m} \sim U(1, |\mathcal{U}_{t, i}|)$.
We choose two $M_{\text{random}} = 16$ and $256$, where former provides a direct comparison 
with our proposed loss estimation approach, and latter represents a brute-force solution that 
works relatively well but is often infeasible in practice due to intensive running time.

\begin{figure}[t]
    \centering
    \begin{subfigure}[h]{0.51\linewidth}
        \centering
        \includegraphics[width=\textwidth]{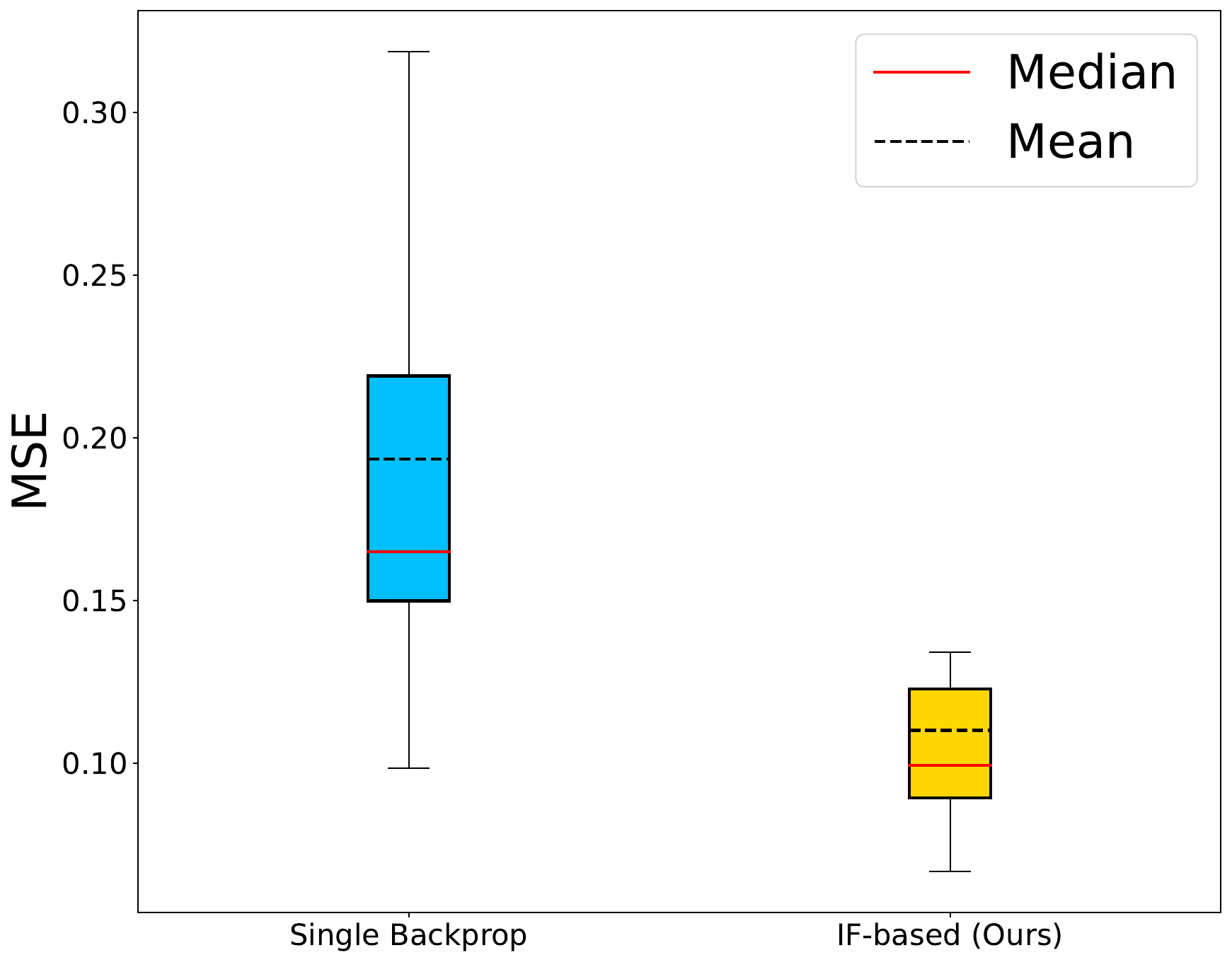}
        \caption{}
        \label{subfig:model_update_mse}
    \end{subfigure}
    \hfill
    \begin{subfigure}[h]{0.477\linewidth}
        \centering
        \includegraphics[width=\textwidth]{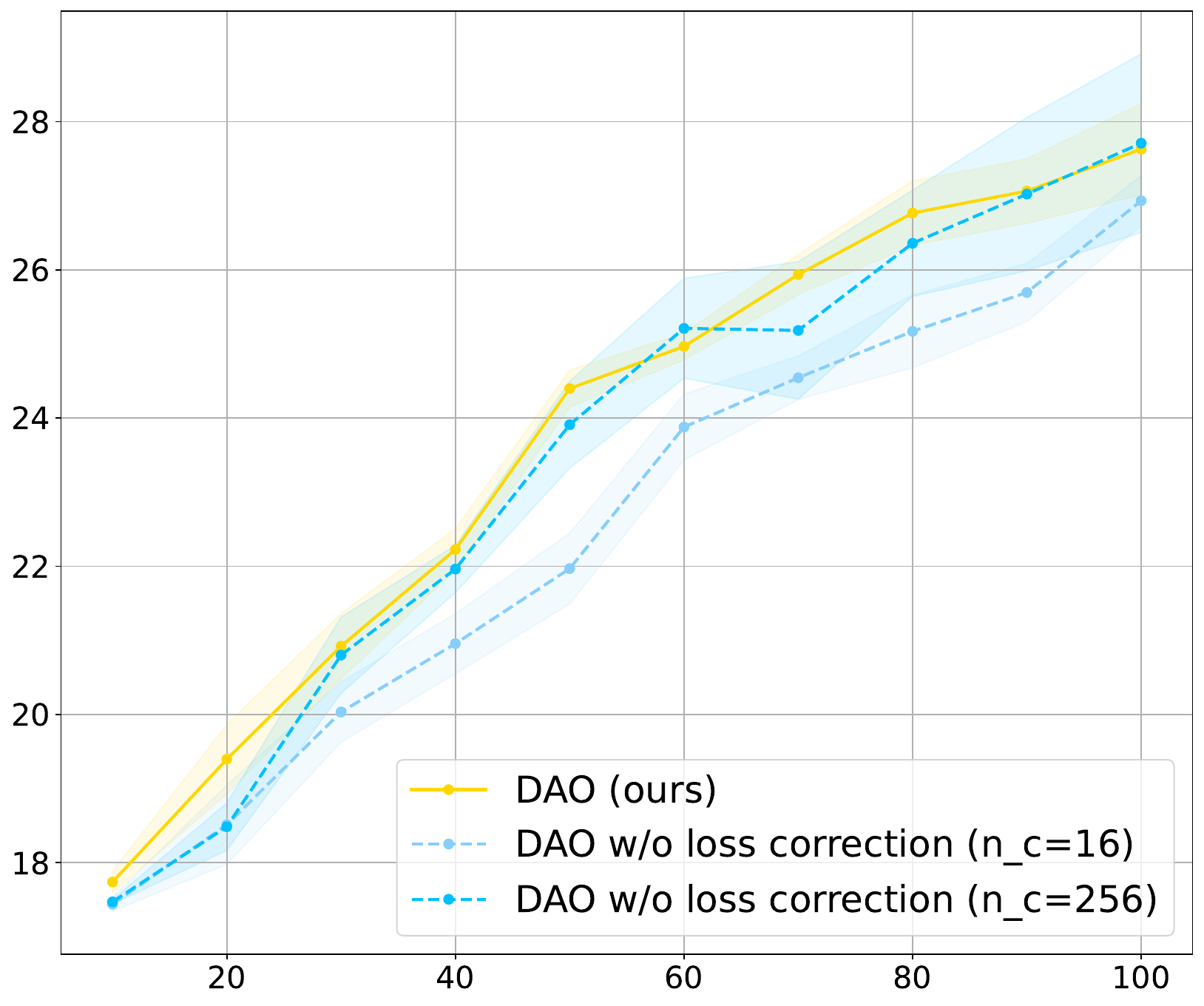}
        \caption{}
        \label{subfig:bias_correction}
    \end{subfigure}
    \caption{
        \textit{Left}: MSE of the predictions accuracy on the test split of CIFAR-10 between 
        models updated by single backpropagation, influence function, and the fully trained 
        model.
        \textit{Right}: Ablation results where the proposed loss estimation is replaced by
        the random sampling estimation defined in \secref{subsec:bias_vs_random}.
    }
    \label{fig:component_analysis}
    \vspace{-0.1in}
\end{figure}

The results are showed in \figref{subfig:bias_correction}.
We see that the proposed method performs even better than the conventional random-sampling loss 
estimation with large sampling size, while computationally being only 1/8 of the run time.
Additionally, the variance of our method is much smaller, indicating more robust loss
estimation, which translates to more robust acquisition performance.
%

\section{Related Work}\label{sec:related_work}
\paragraph{Active learning.}\hspace{2pt} 
AL has gained a lot of attraction in recent years, with its goal to 
achieve better model performance with fewer training data~\cite{freund1997selective, graepel2000kernel, schohn2000less, fine2002query, hoi2006batch, guillory2011online, yang2013buy, chen2021knowledge}. 
There have been different selection criteria including uncertainty~\cite{lewis1995sequential}, 
query-by-committee~\cite{seung1992query}, version space~\cite{mitchell1982generalization} and 
information-theoretic heuristics~\citep{hoi2006batch}. 
The uncertainty-based selection criteria is arguably the most popular, which includes methods 
like least confidence~\cite{lewis1995sequential}, minimum margin~\cite{scheffer2001active}, and
maximum entropy~\cite{settles2009active}.
At their core, these methods select points where the classifier is least certain. 
However, uncertainty-based methods can be biased towards the current learner. 
Query-by-committee and version space~\cite{seung1992query, abe1998query, mitchell1982generalization}, on the other hand, keep a pool of models, and then select samples 
that maximize the disagreements between them. 
Information-theoretic methods~\cite{hoi2006batch, barz2018information} typically utilizes 
mutual information as criteria. 

\paragraph{EER-based acquisition criterion.}\hspace{2pt} 
Alternatively, 
EER was proposed to select new training 
examples that result in the lowest expected error on future test examples, which directly 
optimizes the metric by which the model will be evaluated~\cite{roy2001toward}.
In essence, EER employs sample selection based on the estimated impact of adding a new 
data point to the training set, rather than evaluating performance against a separate 
validation set, meaning that it does not inherently require a validation hold-out set.
However, its necessity to retrain the model for every possible candidate sample and
every possible label renders its cost intractable in the context of deep neural 
networks~\cite{budd2021survey, vskoda2020active, ding2023active}.
More recent EER-based AL algorithms~\cite{killamsetty2021glister, mussmann2022active} 
focus on addressing this efficiency concern.
However, these methods rely on a small set of 
validation data to be used for the evaluation of the expected loss reduction, which is
not ideal for the low-budget AL settings.
In this paper, we present \algname, a novel AL algorithm that improves upon EER through 
optimizations on both model updates as well as loss estimation, efficiently and 
effectively broadening the applicability of EER-based algorithm.
%
%

\section{Conclusions and Future Work}\label{sec:conclusion}
In this paper, we introduced Direct Acquisition Optimization (\algname), a novel algorithm 
designed to optimize sample selections in low-budget settings. 
DAO hinges on the utilization of influence functions for model parameter updates and a separate 
acquisition strategy to mitigate bias in loss estimation, represents a significant optimization 
of the EER method and its modern follow-ups. 
Through empirical studies, DAO has demonstrated superior performance in low-budget settings, 
outperforming existing state-of-the-art methods by a significant margin across seven datasets.

Looking ahead, several promising directions for future research can be explored. 
First, further exploration into the scalability of \algname in larger and more complex datasets 
will be crucial.
%
%
Second, an in-depth investigation into the influence function's behavior in different model 
architectures could yield insights that further refine and enhance the \algname framework. 
%
%
And finally, integrating \algname with other machine learning paradigms, such as unsupervised 
and semi-supervised learning, could lead to the development of more robust and versatile active 
learning frameworks. 

\section*{Acknowledgement}
This work was supported in part by NSF
RI:2313130, NSF 2037026, the Data Science Institute's AI+Science Research Initiative, and the Research Computing Center at the University of Chicago. Any opinions, findings, conclusions, or recommendations expressed in this material are those of the authors and do not necessarily reflect the views of
any funding agencies.
\bibliography{references}
\bibliographystyle{icml2024}

\newpage
\appendix
\onecolumn
\section{Appendix}\label{sec:appendix}
\subsection{More Empirical Analysis Results}\label{subsec:lineplot}
\begin{figure}[H]
    \centering
    \includegraphics[width=\linewidth]{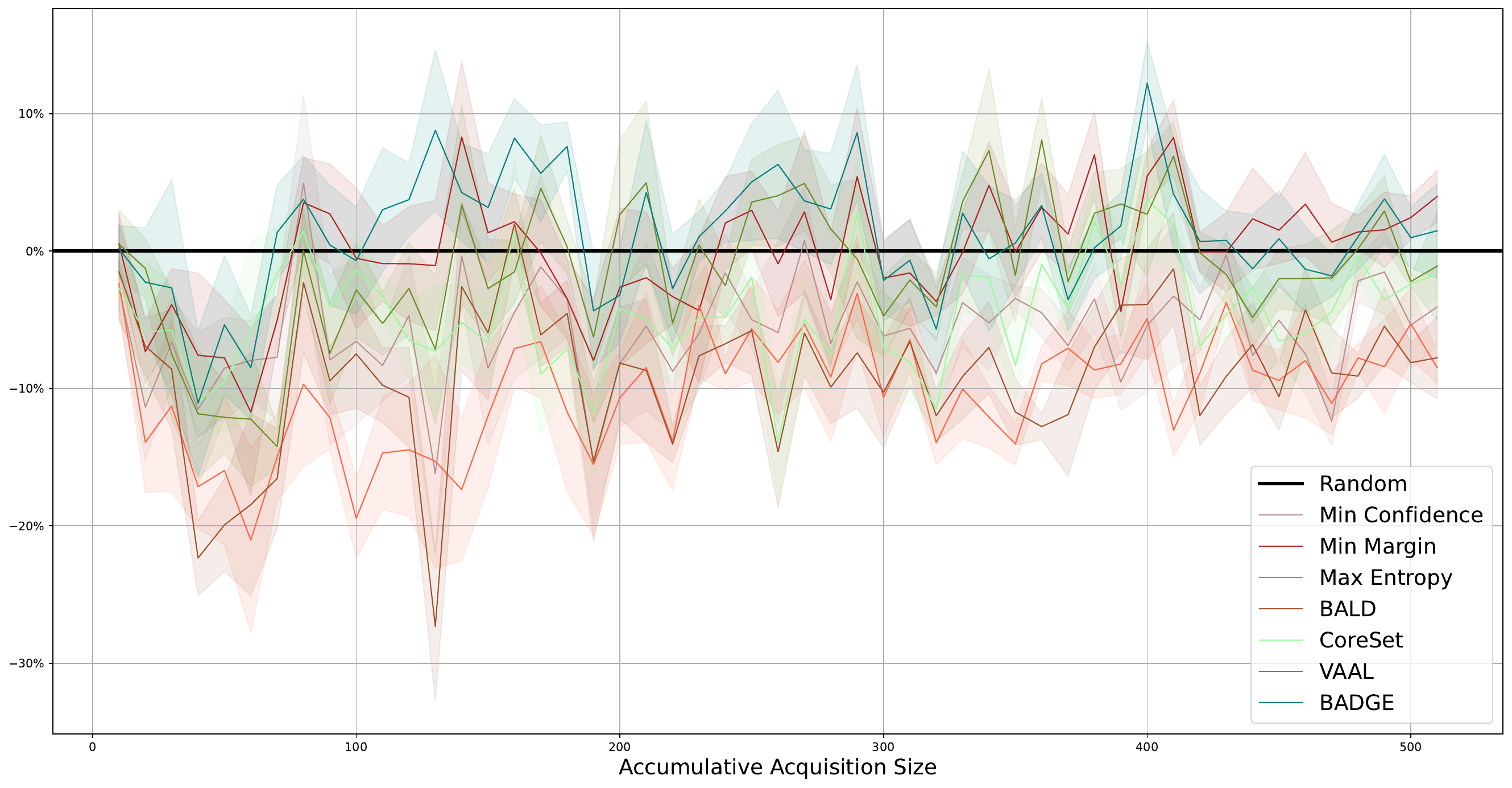}
    \caption{
        Relative performance between existing popular AL methods and random acquisition.
        horizontal axis represents the accumulative size of the labeled set, 
        while vertical axis indicates relative performance in percentage.
    }
    \label{fig:motivation_lineplot}
\end{figure}

\subsection{Experiments on Different Stochastic Batch Acquisitions}\label{subsec:appendix_batch}
\begin{figure*}[h]
    \centering
    \includegraphics[width=0.6\textwidth]{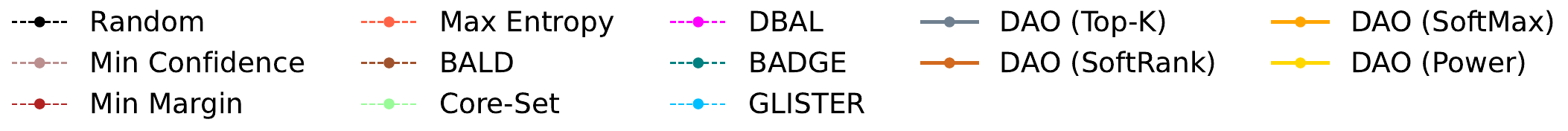}\\
    \begin{subfigure}[h]{0.24\textwidth}
        \centering
        \includegraphics[width=\textwidth]{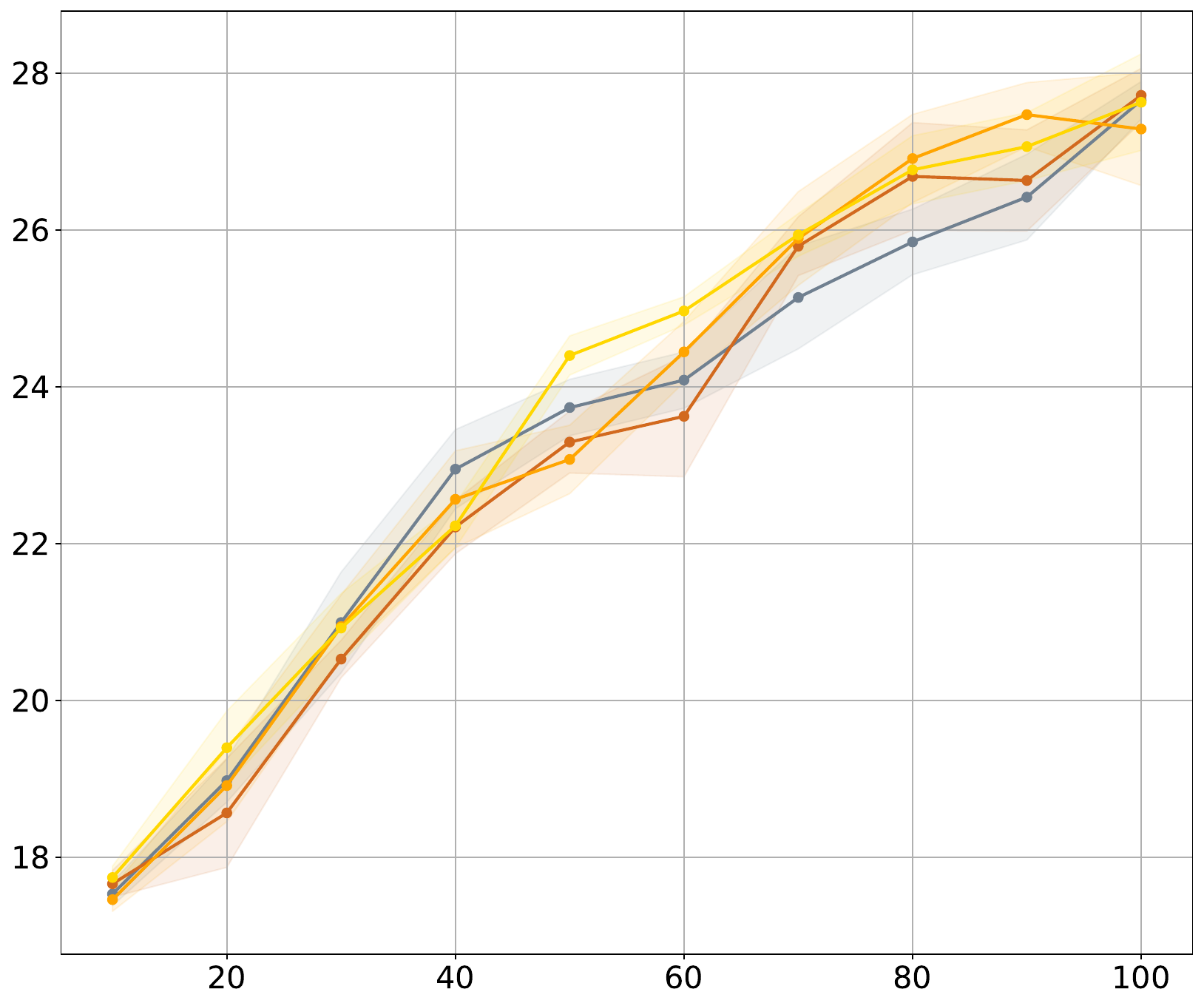}
        \caption{}
    \end{subfigure}
    \begin{subfigure}[h]{0.24\textwidth}
        \centering
        \includegraphics[width=\textwidth]{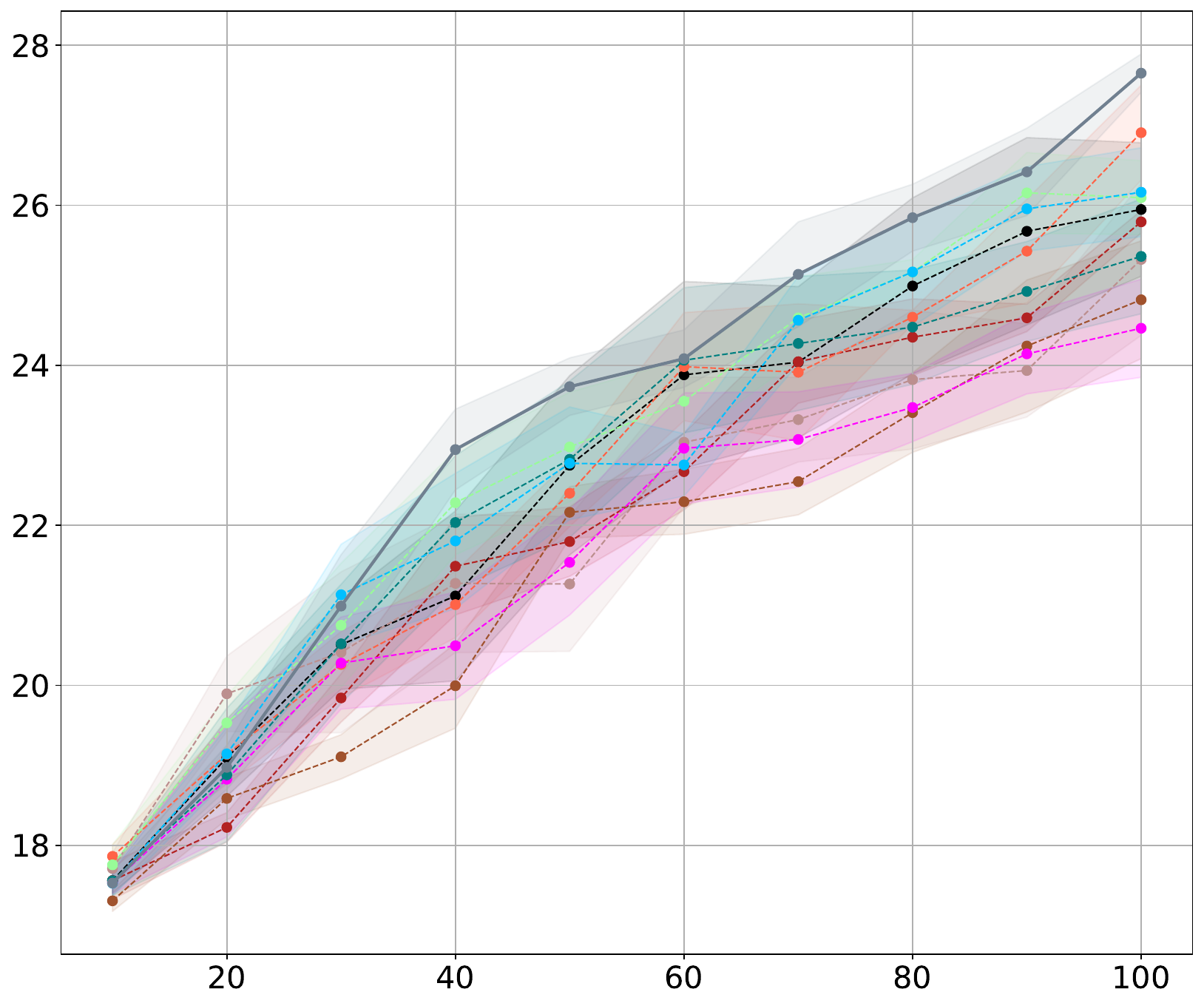}
        \caption{}
    \end{subfigure}
    \begin{subfigure}[h]{0.24\textwidth}
        \centering
        \includegraphics[width=\textwidth]{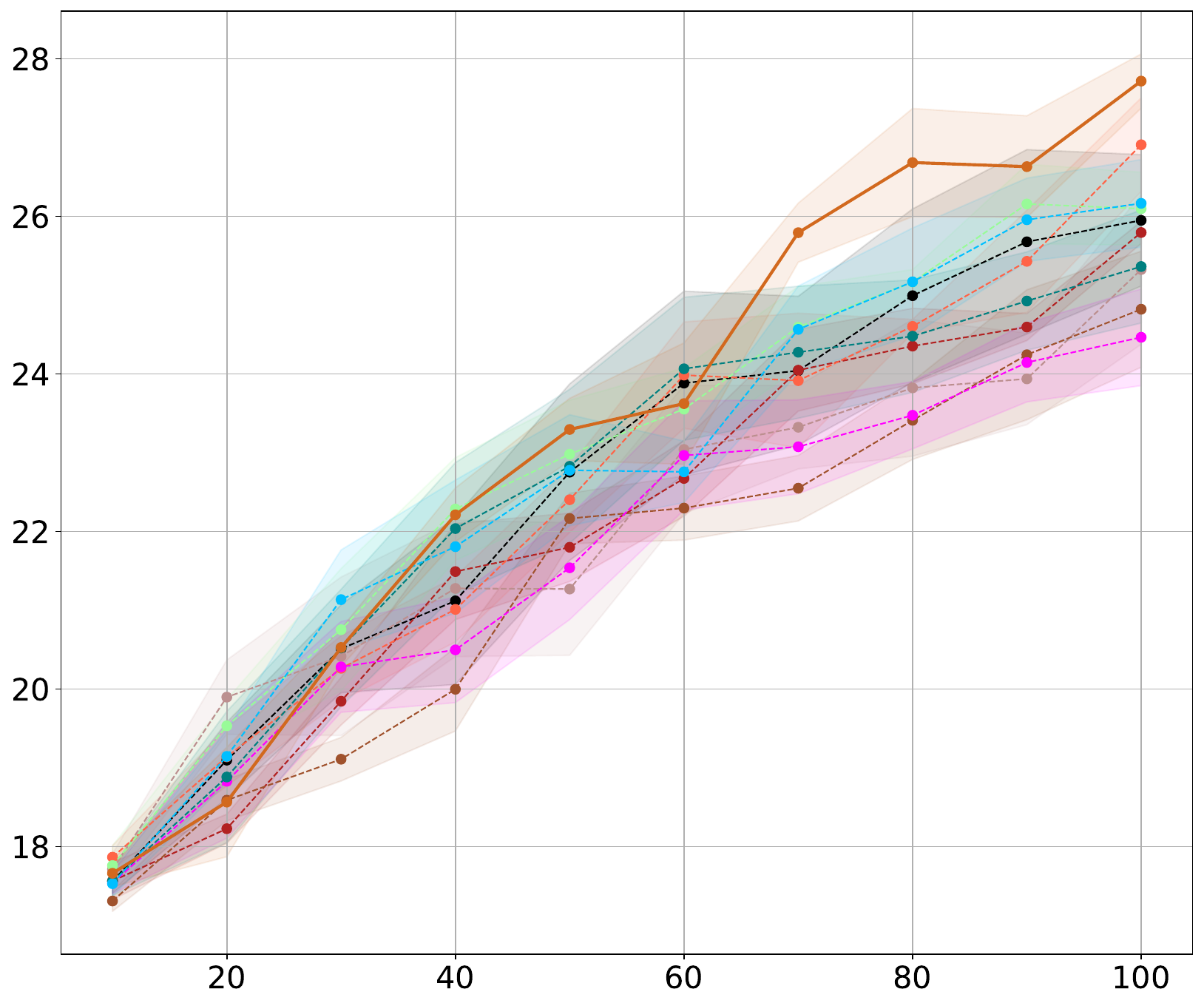}
        \caption{}
    \end{subfigure}
    \begin{subfigure}[h]{0.24\textwidth}
        \centering
        \includegraphics[width=\textwidth]{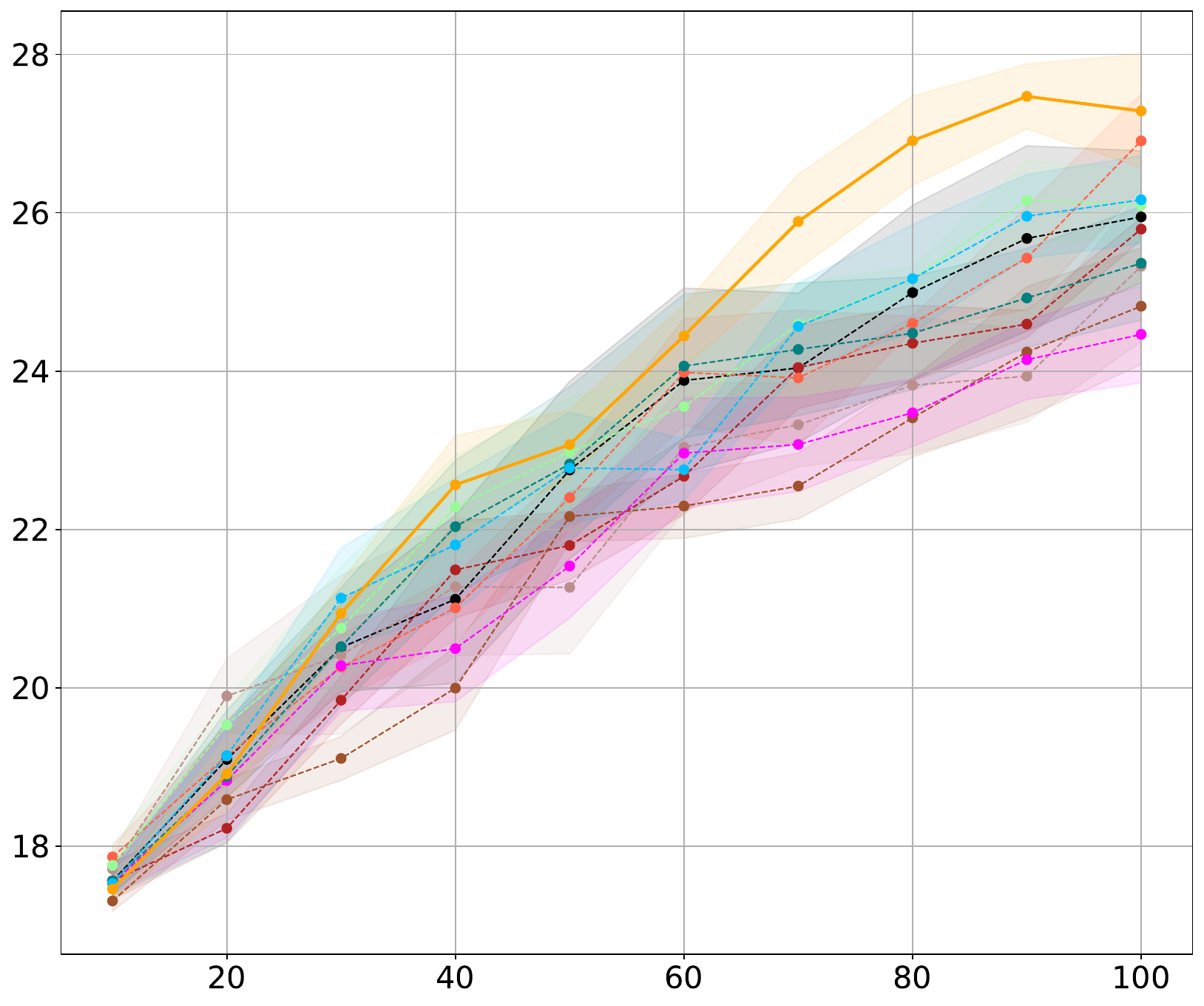}
        \caption{}
    \end{subfigure}
    \caption{
        CIFAR-10 experiment results on
        (a): \algname without batch acquisition strategy (using naive top-k selection) and with
        other sampling strategies (softmax and softrank, as discussed in 
        \secref{subsec:stochastic_sampling});
        (b): \algname without sampling (top-k) vs. existing AL algorithms;
        (c): \algname with softrank sampling vs. existing AL algorithms;
        (d): \algname with softmax sampling vs. existing AL algorithms;
        In all subplots, horizontal axis represents the accumulative size of the labeled set, 
        while vertical axis indicates classification accuracy.
    }
    \label{fig:appendix_results}
\end{figure*}

In this section, we further study the performance of \algname when no batch sampling strategy 
or other sampling strategy is used and compare the results with existing popular AL algorithms.
The results are showed in \figref{fig:appendix_results}.
For all experiments, we used the same low-budget setting as discussed in 
\secref{subsec:object_classification}.
%




\end{document}